\documentclass{article}

\usepackage[english]{babel}
\usepackage{amsmath}
\usepackage{graphicx}
\usepackage{appendix}
\usepackage{array}
\usepackage{multirow}
\usepackage{comment}
\usepackage{caption}
\usepackage{subfig}
\usepackage{booktabs}
\usepackage{authblk}
\newcolumntype{P}[1]{>{\centering\arraybackslash}p{#1}}
\usepackage{lineno,hyperref}
\modulolinenumbers[5]
\providecommand{\keywords}[1]{\textbf{\textit{Keywords---}} #1}

\UseRawInputEncoding

\begin{document}

\title{GPT-3-driven pedagogical agents for training children's curious question-asking skills}

\author[1,2]{Rania Abdelghani\thanks{rania.abdelghani@inria.fr}}
\author[4]{Yen-Hsiang Wang}
\author[3]{Xingdi Yuan}
\author[3]{Tong Wang}
\author[5]{Pauline Lucas}
\author[1,5]{H\'el\`ene Sauz\`eon\thanks{Equal supervision}}
\author[1,3]{Pierre-Yves Oudeyer\thanks{Equal supervision}}
\affil[1]{Inria France}
\affil[2]{EvidenceB, France}
\affil[3]{Microsoft research, Canada}
\affil[4]{National Chung Hsing University, Taiwan}
\affil[5]{University of Bordeaux, France}

\maketitle

\abstract{Children' ability to ask curiosity-driven questions is an important skill that helps improve their learning. For this reason, previous research has explored designing specific exercises to train this skill. Several studies relied on providing semantic and linguistic cues to train them to ask more of such questions (also called \textit{divergent questions}). But despite showing pedagogical efficiency, this method is still limited as it relies on generating the said cues by hand, which can be a very long and costly process.}

In this context, we propose to leverage advances in the natural language processing field (NLP) and investigate the efficiency of using a large language model (LLM) for automating the production of key parts of pedagogical content within a curious question-asking (QA) training. We study generating the said content using the "prompt-based" method that consists of explaining the task to the LLM in natural text. We evaluate the output using human experts annotations and comparisons with hand-generated content. Results suggested indeed the relevance and usefulness of this content.

We then conduct a field study in primary school (75 children aged 9-10), where we evaluate children's QA performance when having this training. We compare 3 types of content : 1) hand-generated content that proposes "closed" cues leading to predefined questions; 2) GPT-3-generated content that proposes the same type of cues; 3) GPT-3-generated content that proposes "open" cues leading to several possible questions. Children were assigned to either one of these groups. Based on human annotations of the questions generated, we see a similar QA performance between the two "closed" trainings (showing the scalability of the approach using GPT-3), and a better one for participants with the "open" training.

These results suggest the efficiency of using LLMs to support children in generating more curious questions, using a natural language prompting approach that affords usability by teachers and other users not specialists of AI techniques. Furthermore, results also show  that open-ended content may be more suitable for training curious question-asking skills.


\section{Introduction}
Curiosity is defined as "the desire to know, to see, or to experience that motivates exploratory behavior directed towards the acquisition of new information" \cite{litman2005}. It is a crucial factor that positively modulates the individuals' learning experience and outcomes~\cite{article,10.3389/fnbeh.2012.00005}. Moreover, research suggests that curiosity is a malleable skill that can be elicited in classrooms by promoting comfort with uncertainty and encouraging questioning to resolve it~\cite{jirout2018, goren2015, law2016, ceha2019a}.

Indeed, relationships between curiosity and questioning have long been studied~\cite{Graesser1994, goupil2023}. For instance, Graesser et al.,~\cite{graesser1992} explore this relation by distinguishing two types of questions : first, convergent (i.e. memory-based) questions that are used to retrieve already-known information~\cite{Graesser1994,Humphries2015} and that are shown to do very little to increase children's curiosity~\cite{bjork2017creating}. Second, divergent questions that are used to gain extra and novel information~\cite{Aschner1963}. These questions require making idea association between two or more facts that are present in the task at hand or linking the task to the individual's previous knowledge. The cognitive process involved in generating divergent questions are shown to be closely related to those involved in epistemic curiosity~\cite{alaimi2020}. However, and despite its importance to curiosity and learning~\cite{jirout2018}, reports show that divergent questioning is almost absent from today's classrooms: children's inquiries are very often low-level questions and memory-based~\cite{Graesser1994,Humphries2015}. Several explanations can be advanced, e.g. children's tendency to overestimate their knowledge and, thus, not having the need to ask further questions~\cite{Graesser1994}, their incapacity to formulate syntactically-correct high-level questions~\cite{Humphries2015}, their fear of their classmates' judgments~\cite{Post2015}, etc. External factors can also lead to influence children's questioning behaviors such as the seating arrangements in the classrooms~\cite{marx1999}, the teachers' domination of the question-asking process~\cite{gall1970}, etc. 

To remedy these problems, previous studies have proposed to take advantage of the benefits of conversational agents (CA) in learning~\cite{Aleven99} and proposed using them to help train children's divergent questioning skills. For example, authors in~\cite{alaimi2020,Abdelghani2022} proposed interactions with CAs that give specific syntactic and semantic cues to help children construct their questions. Their results showed positive effects on question-asking (QA) behaviors as well as on domain-knowledge learning progress. However, the implementation of such technologies is still very limited as it requires a long process of manual generation of the said prompts to control the CA's behaviors. This is, in fact, part of a broader challenge that faces human teachers as well as EdTech designers in general during the production phases. Indeed, having to continuously hand-produce large data-sets of pedagogical content can be a very redundant, time-consuming and a costly task.

In this context, we propose to leverage advances in the natural language processing (NLP) field and investigate the efficiency of using a large language model (LLM) to partially automate the production of pedagogical content for a specific pedagogical task, i.e. divergent QA training. More particularly, we use the GPT-3 language model with a method called "prompting" that consists of describing a task in natural language to LLMs in order to generate a specific output \cite{Brown2020LanguageMA}. In our case, this output represents specific linguistic and semantic cues that can help children formulate divergent questions about a task they have at hand. We investigate two different approaches to generate these cues and evaluate their quality using human expert annotations and comparisons with hand-generated content. We also conduct a field study (75 children aged 9-10) and use human annotations to evaluate children's divergent QA performance when having this training with hand-generated content vs. GPT-3-generated one. This investigation study is crucial as LLMs can still show poor performances despite their impressive powers~\cite{kasneci2023, bender2021} e.g. reasoning and cognitive biases, hallucination problems, etc.

Our study contributes to the ongoing effort of unlocking the challenges of using LLMs in educational applications in general and in curiosity trainings in particular.

\section{Related work}
\subsection{Epistemic curiosity and QA training in education}
Curiosity plays an important role in learning from infancy~\cite{stahl15} to adulthood~\cite{reio2000}, and in fostering academic achievement, particularly for children with low socio-economic status ~\cite{prachi2018}. It can be seen as a a personality trait (\cite{kashdan2004}), a psycho-emotional state aroused by external situations(\cite{mehta2018}, \cite{kang2009}),\cite{berlyne1954},...) or as a malleable skill that can be trained through teaching specific practices. Indeed, several theories and frameworks such as Lowenstein's "Information gap theory" (\cite{loewenstein1994} or Murayama's process account of curiosity in learning (\cite{Murayama2019}), suggest that enhancing one's curiosity-driven learning should focus on training the awareness of missing information and the appropriate inquiry strategies to compensate for it. 

One of the most important inquiry strategies to be trained is question-asking. More particularly, research such as in~\cite{alaimi2020} suggest that the more curious students are by trait, the more divergent-thinking questions they will ask, i.e., high-level questions that require making hypotheses, predictions, ideas inferences, and that bring additional information~\cite{Aschner1963}. This includes for example questions that involve guessing the cause of an event, predicting the consequences of an event, or that help make sense of the task and explore it in greater details. 

In this context, Abdelghani et al.,~\cite{Abdelghani2022} explored using CAs to train children's divergent QA skills by proposing specific hand-generated semantic and linguistic cues. The study then investigated whether such trainings can help children engage in intrinsically-motivated explorations of their learning environments and gain new knowledge on their own. Their results showed, indeed, positive effects on the curiosity-driven behaviors and subsequent domain-knowledge learning progress. However, this study presented important limitations since the agents had very low intelligence level (i.e. guided by hand-generated cues) and restrained children to think about specific and predefined questions.

\subsection{Natural language processing and conversational agents for building educational applications}
Advances in the NLP field have helped create several opportunities to build innovative and high-quality educational applications. 
Work in this area has facilitated the semantic, syntactic and grammatical analysis of students' typed and spoken inputs and the detection of their potential errors. Another interesting use of NLP in education is the automation of tasks that are traditionally executed by hand such as creating the curriculum and the evaluation materials content. Indeed, authors in \cite{Sultan2014} for instance, showed that NLP-based methods can be used to extract the key scientific ideas and concepts from science educational resources that, according to human experts, are important for a better and deeper understanding of the domain. More importantly to this study, NLP-based methods have also shown to be efficient in the question-generation tasks (i.e. generating questions from educational resources) which can help teachers alleviate the time-consuming process of creating the assessment and evaluation materials (\cite{Brown2005}, \cite{Steuer2021}, ..).

On a final note, it is also important to say that earlier research has showed several positive pedagogical impacts of using NLP-based social conversational agents on children's perception of their own competencies and their attitudes towards learning (\cite{Silvervarg2010}). 

\subsection{GPT-3 and prompt-based learning}
Over the past few years, the NLP field has been focusing on the use of pre-trained large language models (LLMs) as they demonstrate impressive powers in generating natural language and are easier to use in downstream tasks (\cite{devlin2019,Brown2020LanguageMA}). Indeed, LLMs are trained on enormous amounts of text data, giving them the ability to learn novel tasks only by taking a few examples as context. 

However, adapting a pre-trained LLM to the context of a specific downstream task is still a quite complex task as it mostly relies on fine-tuning: a learning method that requires performing a further training of the pre-trained parameters of the LLM using labelled data that is specific to the task in question. This method can be problematic for several reasons: it can be extremely costly and hard to perform for super large models that have an enormous number of parameters~\cite{Raffel2020}, data can be scarce in many cases, the model is fine-tuned separately for each different task, etc. As an alternative, methods such as prefix-tuning have emerged~\cite{li-liang-2021-prefix}, where only a small percentage of the model's parameters are updated for task-adaptation. We also find adapter-tuning methods~\cite{https://doi.org/10.48550/arxiv.2106.03164} that only add a few trainable parameters per new task, allowing a high degree of parameter sharing. These simpler techniques have achieved comparable results to fine-tuning but still require machine learning expertise for their implementation, which can be a major brake for using them in educational settings where most stakeholders do not have this expertise.

In this context, GPT-3, one of the recent and most performing LLMs, came along with an even simpler technique called "prompt-based learning": it is a new way to handle a wide range of NLP tasks by leveraging the LLM's ability to model natural language. It consists of verbalizing the task to be completed in natural text (this is what is called a "prompt"), with no update of the model's parameters. It is also possible to provide examples in the "prompt" to show the model how the task should be completed; this is known as K-shot learning, where K refers to the number of examples provided. This technique has been applied in a wide range of applications such as knowledge-based question-answering~\cite{jiang-etal-2021-know} but, up to our knowledge has not yet been applied for curiosity-based questions generation. In addition, prompt-based learning with GPT-3 has achieved very strong performances in zero-shot and one-shot settings (i.e. with only zero or one labelled example), without any fine-tuning~\cite{Brown2020LanguageMA}. This is an enormous advantage for this study as it allows to implement a simple-to-use technology that can be easily usable and/or adapted for a different context by the education community (i.e. teachers, caregivers ..) without having to have any expertise in machine learning.

\section{Current study}
Here, we aim to study the efficiency of using GPT-3 for generating linguistic and semantic cues that can help children formulate divergent questions. We compare this approach with hand-designed cues using human experts annotations of: 1) the generated cues themselves; 2) their learning outcome in children in a field study. Concretely, we proceed like the following :

\begin{itemize}
    \item In a first step, we use GPT-3 to generate curiosity-eliciting cues and compare them to what human experts can produce when given the same task. We test generating two types of cues : 1) incentive 'closed' cues that are composed of a short sentence to guide children towards thinking about one specific relevant information they might be missing about the task at hand + a questioning word that they can use in order to formulate a divergent question concerning this identified missing information (this type of cues was previously studied in \cite{Abdelghani2022}. 2) 'open' cues that can lead children to pick between several missing information and questions they can pursue + a list of questioning words they can use. We then use human experts annotations and comparisons with hand-generated content in order to evaluate the output of GPT-3 in terms of semantic relevance, divergence level and syntactic quality.
    \item In a second step, we conduct a field study (75 children aged 9-10 in primary school) to evaluate children's divergent QA performance when having these different kinds of cues. To do this, we also use human experts annotations to compare the questions they formulated, in terms of divergence level and overall quality, when interacting with a CA that proposes incentive 'closed' cues generated by hand (incentive hand-crafted CA) vs. generated by GPT-3 (incentive GPT-3 CA) vs. when having 'open' cues (open GPT-3 CA).
    \item Finally, we evaluate the training's effect on children post the 3-day intervention, using specific tests. These tests include children's learning progress with respect to their ability of asking divergent questions, their perception of their own competency in this skill and their attitude towards curiosity and asking questions in general.
\end{itemize}

\section{Study design}
"Kids Ask" is a prototype for a web-based educational platform that involves an interaction between a child and a CA during a reading- comprehension task. The platform was designed during our last study~\cite{Abdelghani2022} when it only included hand-crafted CAs. It is designed to propose a curiosity training focusing on divergent question-asking. It offers 2 workspaces : one to stimulate curiosity via encouraging self-reflection (workspace 1) and one to train divergent QA skills using the cues proposed by the CA (workspace 2).

In order to answer the aims of the current study we described above, we use two different implementations for the CA in the workspace 2 : it proposes cues that were either generated by hand or by GPT-3 ; see more details about these different implementations in the experimental conditions section~\ref{section:exp-conditions} and their technical implementation in Section~\ref{section:tech-impl}. 

\begin{figure}
\centering
\captionsetup{justification=centering,margin=1cm}
\fbox{
\includegraphics[width=.7\linewidth]{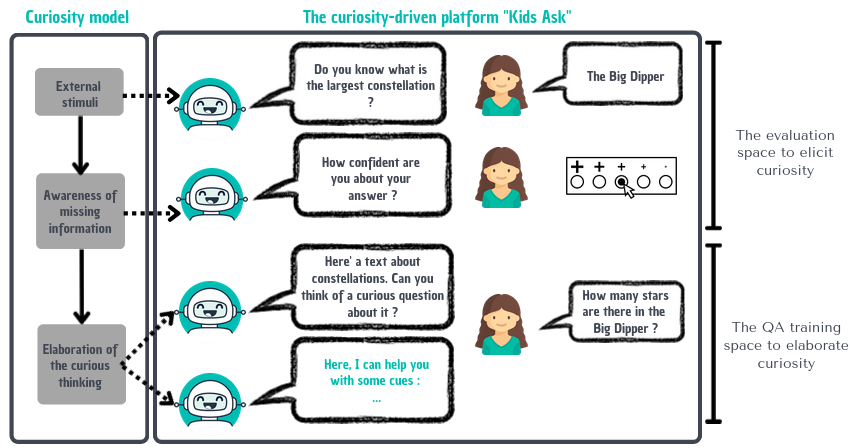}}
\caption{\textbf{Illustration of the agent's strategies in the different work spaces of "Kids Ask", with relation to our curiosity model}}
  \label{fig:pres-ka}
\end{figure}

Concretely, the two workspaces are the following (see Figure\ref{fig:pres-ka}) :
\begin{itemize}
    \item \textbf{Workspace to elicit curiosity via encouraging self-reflection}: This space contains a general knowledge quiz; the items can be skipped if children decide that they don't know the answer. However, if they submit an answer, the agent asks them to report a confidence level in this answer with a 5-Likert scale (from "Super not confident" to "Super confident"). This strategy is inspired from studies such as \cite{Graesser1994} where authors suggest that allowing children to skip questions and to report states of their own knowledge can help them identify more uncertainties which, in its turn, may motivate subsequent curiosity-driven behaviors such as divergent QA~\cite{Roebers2007, berlyne1954}.
    
    \item \textbf{Workspace to train question-asking skills (The QA training space)}: In this space, children have different texts relating to a theme of their choice: they choose one of the six themes that they have worked on in the first workspace. For each text, they interact with the agent that will try to help them think of divergent questions by giving them specific linguistic and semantic cues. These cues are a questioning word that is either combined with 1) an answer to a specific question about the text, that was generated by hand, 2) an answer to a specific question about the text, that was generated by GPT-3, or 3) important key words that are related to the text and that were generated by GPT-3. See the Experimental Conditions Section \ref{section:exp-conditions} for more details about the composition of these cues. 
    
    The agent chooses to give one of these three types of cues depending on the condition the child is assigned to: hand-crafted incentive agent, GPT-3-driven incentive agent or GPT-3-driven open agent. For the three conditions, and similarly to the previous study in ~\cite{Abdelghani2022},the agent does not give any feedback concerning the question formulated by the child.
    
Participants started with the general quiz relating to 6 themes using the first workspace. They then chose their favourite topic and moved to the QA-training space where they had reading tasks related to their topic of choice.
\end{itemize}

\subsection{Experimental conditions}\label{section:exp-conditions}
Our aim is to evaluate GPT-3's efficiency in generating cues that can elicit divergent QA in children, compared to human experts. We also use this study to test 2 different types of cues and investigate the difference of their impact on children's performance. To meet this aim, we set up three experimental conditions :
\begin{enumerate}
\item Hand-crafted incentive agent : A control group where the agent is controlled by human-generated incentive cues that lead to predefined specific divergent-thinking questions, i.e., high-level questions that require making hypotheses, predictions, ideas inferences... \cite{Aschner1963}) (Group 1). 
\item Automated incentive agent : A first test group where the agent is controlled by GPT-3-generated incentive cues that lead to predefined specific divergent-thinking questions (Group 2). 
\item Automated open agent : A second test group where the agent is controlled by GPT-3-generated open cues that lead to several possible divergent-thinking questions about the text (Group 3).
\end{enumerate}

More concretely, the cues for groups 1 and 2 consist of : a questioning word + an answer to one possible divergent question about the text. Combined, these two cues are meant to lead children to think of one specific divergent question about the text. \textbf{Example:} For a text about Louis Pasteur, the agent gives the cues \textbf{'What difference'} and \textbf{'The vaccine avoids the disease whilst medicine treats it'} in order to lead the child to ask the question \textbf{'What is the difference between vaccines and medicine?'}. This question is considered to be divergent as its answer is not explicitly available in the text and requires the child to link two ideas~\cite{Aschner1963}. For the Group 1, these cues were chosen by researchers and teachers as they were judged to be leading to important questions for children to ask. For the Group 2, these cues were generated automatically by GPT-3 (see details about the technical implementation in section~\ref{section:tech-impl}).

For the third group, the agent proposes two key words that are closely related to the key ideas of the text. These words constitute cues that can lead to think about several different divergent questions about the text. To be in the same conditions as the other groups, we also give a possible questioning word. But because the possibilities are multiple in this condition, the agent tells children that they can use different starters if they want to and reminds them of the most commonly-used ones. \textbf{Example:} For a text about Louis Pasteur, the agent gives the cues \textbf{What other} and \textbf{"Canning, Freezing"}. Multiple divergent questions are possible with these words: \textbf{"What is the difference between canning and freezing ?", "Why is Pasteurisation better than freezing ?", "How does canning work ?"...}.

\begin{figure}
\captionsetup{justification=centering, margin=1cm}
  \centering
  \fbox{
  \includegraphics[width=.7\linewidth]{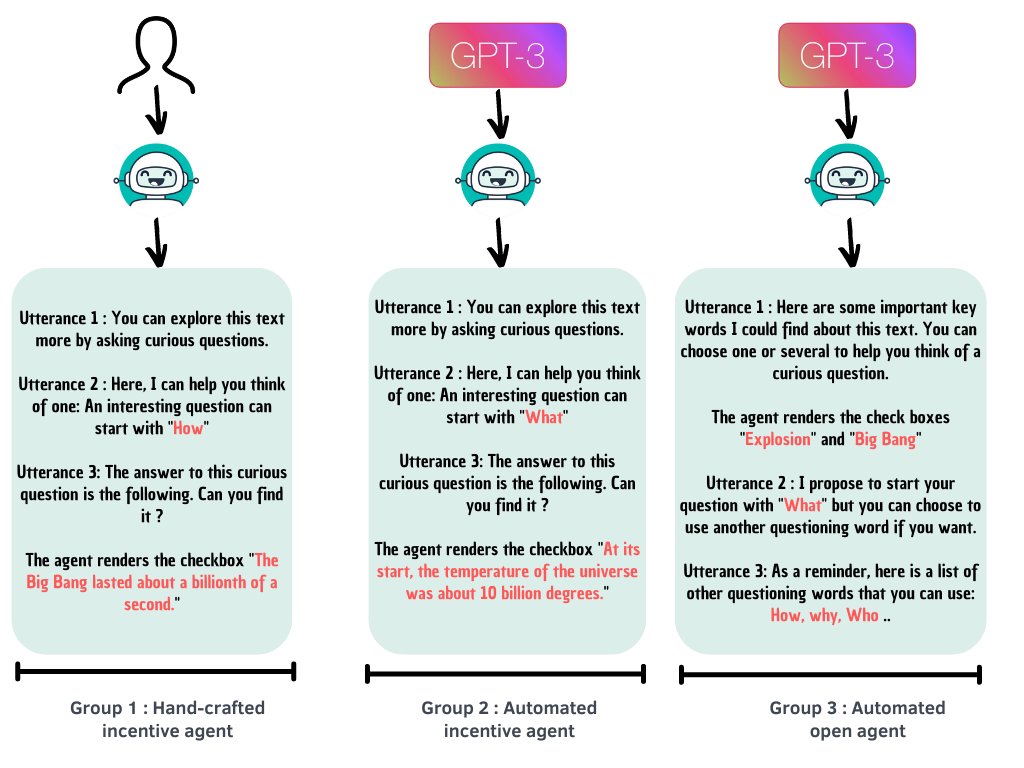}}
  \caption{\textbf{Difference between the agents' behaviors and implementations for the three experimental conditions}}
  \label{fig:conditions}
\end{figure}

See Section~\ref{section:choice-cues} for the elaborate rationale behind the choice of these cues, Figure \ref{fig:conditions} for details about the difference between the agents' dialogues and implementations for the different conditions and Section~\ref{section:procedure} for snips of the agents' behaviors in "Kids Ask".

\subsection{Design rationale}
In this section, we explain the choices we made for the design of our study. More particularly, we discuss our choice of the interaction flow, our choice of the cues proposed by the CA to elicit divergent questioning and finally, our choice of using the GPT-3 LLM with prompting.

\subsubsection{Choice of the interaction workflow}
As mentioned above, we choose to start our interaction with a skippable general knowledge quiz in order to elicit children's curiosity about one or different topics that we propose. Since this quiz contains items relating to different topics, it also serves us as a tool to help children choose the topic they want to work on during the next task, in the QA training space. Indeed, giving children the choice over their tasks has shown to have a strong positive effect not only on their motivation and engagement but also on the amount of knowledge they can learn and their perceived competence~\cite{Cordova1996}.

\subsubsection{Choice of the cues proposed by the CAs}
\label{section:choice-cues}
For the QA-training space, we focus on divergent questions in general as they require higher-level thinking and are more related to epistemic curiosity (\cite{Aschner1963, Bereiter1992}). In order to guide children towards this type of questioning, we choose specific cues for our agents to propose (as described in \ref{section:exp-conditions}). This choice is based on three main ideas: 
\begin{enumerate}
    \item The first type, i.e. the answers to specific divergent questions that the agents give to Group 1 and 2, are motivated by the results of two previous studies \cite{Abdelghani2022, alaimi2020} : when generated by hand, this type of cue showed to be rather efficient in enhancing children's divergent QA skills. Therefore, we decide to reuse this cueing strategy and test reproducing it with GPT-3. We evaluate GPT-3's performance using human annotations for 1) the semantic relevance and divergence level of the cues and 2) for children's divergent QA performance when having the hand-generated cues (group 1) vs. GPT-3-generated ones (group 2).
    
    Therefore, and in order to facilitate this agent implementation, we use GPT-3 to generate the exact same type of cues. We then compare children's performances when having the cues generated by humans vs. by GPT-3 (i.e. Group 1 vs. Group 2) in order to investigate the validity of using GPT-3 for this task. 

    \item The motivation behind the second type of semantic cues (i.e. two keywords that the CA gives to Group 3) is to propose cues that leave children with more choice over the questions they can formulate during the training. Indeed, having heavily teacher-directed tasks that impose specific tasks to all children (like the case for our first type of cues) may alter with their sense of control and autonomy which are central factors for motivation and curiosity according to the Self-Determination Theory(~\cite{ryan2000selfdetermination}, \cite{Guay2022}).
    
    Furthermore, proposing the same pieces of information for all children is likely to result in giving trainings that do not match with their different competency levels and zones of proximal development (ZPD). Therefore, leading children to ask specific questions can lead them to ask questions that are too complicated for them or, on the contrary, questions that they already know the answer to. This can mean that the training is likely not to support the users' curiosity since this latter is tightly related to each individual's ZPD and sense of competency~\cite{METCALFE202040}. Finally, we also think that it is important to give different choices in order to avoid the cases where teachers' or researchers' choice of the semantic cues is heavily influenced by their preconceptions and beliefs about children's knowledge gaps~\cite{john1998, tullis2015}.
    
    Therefore, we propose propose to use GPT-3 to extract relevant keywords about a given text. These keywords can be used by the CA to raise children's awareness about the important concepts of the texts that need further investigations, and guide them towards thinking of different divergent questions. Investigating children's divergent QA performance when having these GPT-3-generated open vs. specific cues (i.e. Group 2 vs. Group 3) could therefore help us investigate different contexts where children's curiosity can be induced more efficiently. 

    \item Finally, we choose to maintain the linguistic help (i.e. the questioning word that can be used to start the question) for our three conditions as question-generation tasks can be quite difficult for children, as shown in \cite{Graesser1994}.
\end{enumerate}

\subsubsection{Choice of the technology}
We choose to work with the pre-trained LLM GPT-3 and to use prompt-based methods\footnote{It is to be noted that in this article we use the word "prompt" in two different contexts : (1) prompts we give GPT-3 to generate the cues proposed by the CA and (2) prompts we give to children to help them generate divergent questions.} because of two reasons. First, we choose to work with GPT-3 given that it has significantly more parameters and was trained upon text corpus containing a wider variety of topics, compared to its predecessor GPT-2 for example. Moreover, it has demonstrated robust performances in various downstream tasks, and more importantly to this study, in knowledge-based question-asking and question-answering tasks~\cite{Xingdi2022}. Indeed, such results motivate a follow-up work to move towards more complex types of questions like high-level and divergent questions. 

Second, and with the aim to implement an easy-to-use system that can be accessible for the broad teaching community, we choose to work with GPT-3 in a prompt-based methods setting (\cite{Brown2020LanguageMA}). Indeed, and as discussed above, prompt-based learning consists simply of giving pieces of natural text to the model as an instruction in order to adjust it to the specificities of the task in question and generate the output needed (in our case, this output is the set of cues as described in the Conditions section \ref{section:exp-conditions}). So with this simple implementation, we avoid complex methods such as fine-tuning that require a machine learning expertise and collecting a large set of specific data, etc. Instead, we provide a system that can be easily used by practitioners (e.g. teachers) and/or adapted to different activities of their choice (e.g. change the language, target a different type of questions ..), without the need for a specific expertise.

\subsection{Ethical considerations}
One of the first challenges we faced during the design of our LLM-based system is ensuring safe interactions with children. Indeed, despite the impressive advances, LLMs are still considered as 'black-box' systems that are impossible for us to control and that can generate stereotypes and wrong/ biased content~\cite{bender2021, blodgett2021}. For this reason, we choose to work with a human-in-the-loop setup aiming to have human evaluators verify the content generated by the LLM before proposing it to children. We also chose to use such an offline configuration (i.e. not having the model connected directly to the web application) in order to ensure the privacy of children's data and avoid having their personal data retrieved by the LLM. (see details about this implementation in Section~\ref{section:tech-impl})

Another challenge we face with our method is children becoming excessively reliant on such LLM-based systems. This can indeed affect their problem-solving and creative skills~\cite{kasneci2023}, as well as their 'real-life' interactions. To address this, it is very important to accompany the use of such systems in schools with pedagogical interventions that aim to raise awareness about the limitations of LLMs. This can help children develop critical-thinking when faced with LLM-generated content and see that it cannot replace other sources of information like books, interaction with classmates, teachers, etc. It is also very important to note that the use of such systems should be completely monitored by the teacher to respond to their pedagogical need (i.e. frequency of use, collaborative vs. isolated use, etc).

Finally, one other issue that may arise is teachers becoming very reliant on using LLMs in their practices. In this context, it is important to provide reminders that LLMs cannot replace the efficiency of human instructions and that they can only be used as a complement but not as a replacement for human teachers (e.g. to avoid long and redundant tasks like writing several exercises for the same knowledge component, to improve the learning experience and enhance task engagement, etc.)

\subsection{Technical implementation}
\label{section:tech-impl}
\subsubsection{Generating the agents' cues using chain-of-thoughts prompting in GPT-3}
\label{section:prompting-approaches}
\begin{figure}
  \centering
  \fbox{
  \includegraphics[width=.7\linewidth]{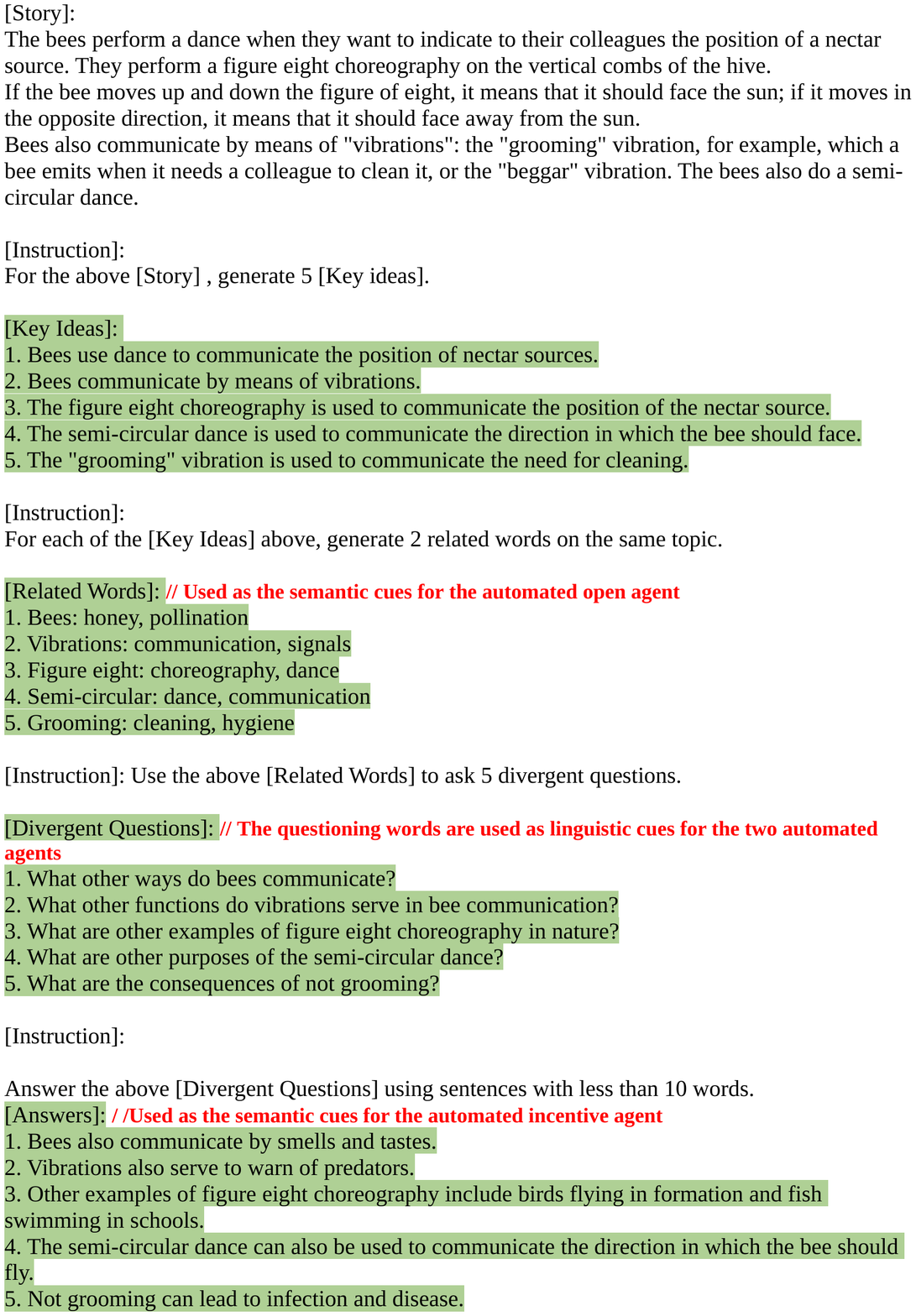}}
  \caption{\textbf{Prompting approaches used to control the agents behaviours}}
  \label{fig:prompting-approach}
\end{figure}

For our experiments, we prompt GPT-3 in a zero-shot setting \cite{Brown2020LanguageMA}, meaning that the model predicts an answer based only on a context and natural language description of the task, i.e. a prompt, without giving it any examples. Concretely, we provide a data-set containing the text-based educational resources: short scientific articles with a mean of 109 words per article, that will serve the model as context. Our task consists of building the proper prompts that can lead GPT-3 to produce a questioning word combined with either a sentence (for the group 2) or a pair of related keywords (for the group 3) that can be used as cues to formulate a divergent question relevant to the context. See Figure \ref{fig:prompting-approach} for details about the prompts approaches we retained.

Before deciding on which prompting approach to choose, we run several trials where the following configurations were tested:
\begin{itemize}
    \item Story then instruction vs. instruction then Story: At the very beginning of the development, we did try a set of different ways to form the prompts, including the ordering of story vs instruction, using special indicators to highlight story/instruction etc. and prompt formats (either to have a new line after the indicator). We do not observe significant difference from these variants, probably because LLM at the scale of GPT-3 is sufficiently robust on this format dimension. We decide to stick with a format (as shown in Figure~\ref{fig:prompting-approach}) that is considerably straightforward and readable so practitioners do not need to specially design the prompts. 
    \item Zero-shot vs. One-shot vs. Few-shots vs Fine-tuning: In all the experiments described in this work, we prompt GPT-3 in a zero-shot manner. The reason is two-fold. 
    \textbf{1)} There are many recent works in the NLP community report that pre-trained LLMs such as GPT-3 can be sensitive to the specific data points provided as few-shot examples \cite{kumar2021,lu2021fantastically,zhao2021calibrate,min2022rethinking}. For instance, LLMs are observed to suffer from ``recency bias'', i.e., they overly rely on the examples that are located closer to the end of the prompt and thus tend to bias the output towards copying from the most recent examples. Despite many works have been proposed to alleviate such issues, it remains to be solved to have a few-shot prompting strategy that is generally working better. 
    \textbf{2)} We want to emphasize that in most real-world scenarios such as pedagogical applications, practitioners are less likely to have machine learning expertise nor sufficient computational resources to fine-tune an LLM, or to optimize the LLM outputs by providing the model with different combinations of example. We envision that the proposed system can have greater accessibility if it does not require too much expertise from the users.
    Lastly, we want to mention that our method is orthogonal and applicable to bleeding-edge prompt-tuning methods from the NLP community \cite{rubin2021learning, lu2021fantastically, liu2022makes}.
\end{itemize}

In all experiments, we use the text-davinci-002 variant of GPT-3 and a temperature of 0.7 during the prompting process.

\subsubsection{Agent behavior}
As we have no control over GPT-3's outputs, we choose not to put children in direct contact with the model in order to avoid any potential offensive interactions. To do so, we opt for an offline implementation like the following:
\begin{itemize}
    \item We use GPT-3 to generate the agents' cues like described in the section above.
    \item For more transparency of our proposed method, we repeat the procedure several times for the same resources and we take some of the outputs randomly as cues to appear in the platform (even if they lead to convergent questions).
    \item After verification that these cues do not include any inappropriate or offensive words, we translate them into French using DeepL and include them in a dedicated database. It is to be noted that during all of our experiments, we did not see any offensive output (see more details in the results section~\ref{section:general-res}).
\end{itemize}
The agent is then connected to this database containing the different educational resources and the cues for each one of them, for the three conditions. The cues for the hand-crafted agent are included manually in the database. Depending on the child's assigned condition, the agent composes the dialogue utterances in order to include the appropriate support. We changed the utterances between the questions to avoid repetition in the agent's dialogue: \textbf{Example :} If, for the first question, the agent says \textbf{"Here are two cues to help you think of a curious question about the text, can you use them ?"} than for the following one it will choose another replica such as \textbf{"Can you combine these two cues to generate a curious question about the text ?"} \cite{Jones2018}.

\section{Experimental procedure}
\label{section:procedure}
\begin{figure}
  \centering
  \fbox{
  \includegraphics[width=.9\linewidth]{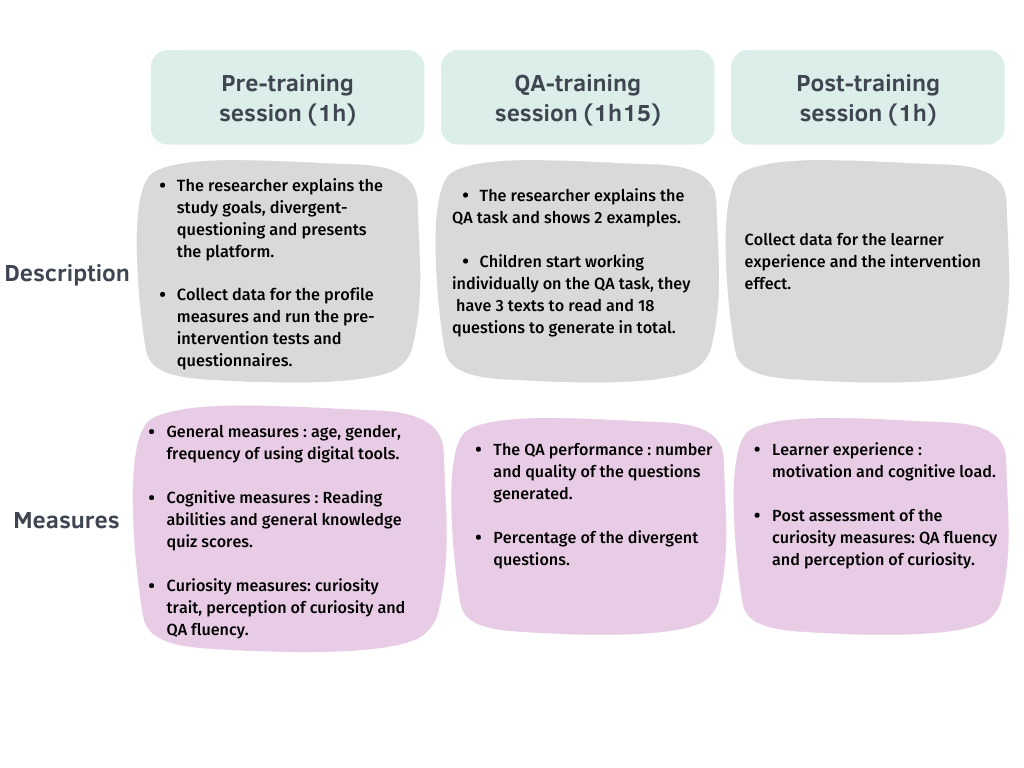}}
  \caption{\textbf{Study timeline and measures}}
  \label{fig:timeline}
\end{figure}

The study consisted of three sessions of 1h within the same week: one for collecting the profile measures and pre-intervention measures mentioned above, one for the divergent-QA training and one for the post-intervention assessment. See study timeline in Figure \ref{fig:timeline} and details of the different measures in the data collection paragraph \ref{section:measures}. 

Our study was approved by the institute's ethics committee (certificate n°2019-23) and started after having all participants parents' signed consents.

\begin{figure}
  \centering
  \fbox{
  \includegraphics[width=.9\linewidth]{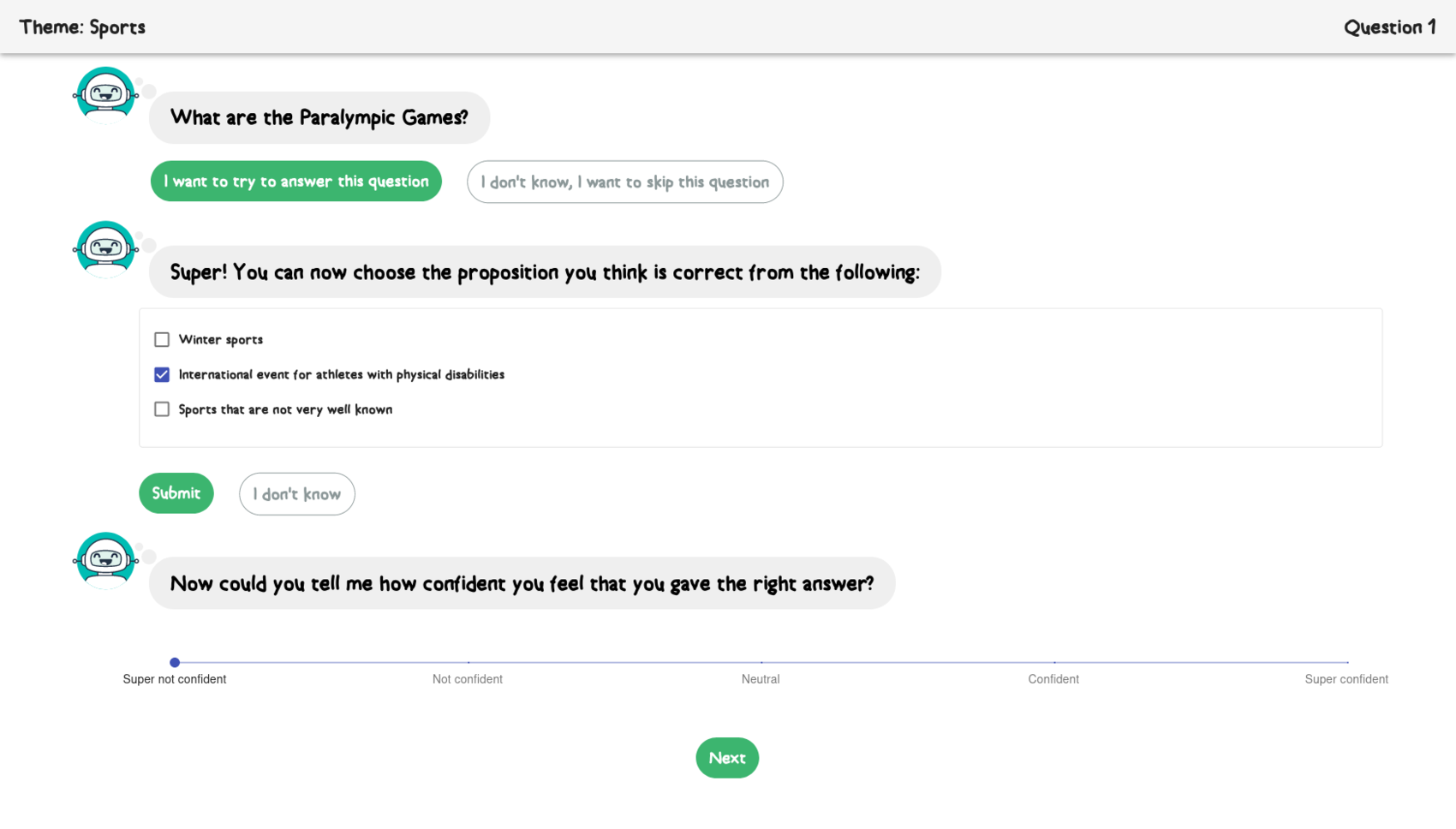}}
  \caption{\textbf{Interface for the domain knowledge quiz}}
  \label{fig:Quiz}
\end{figure}

\textbf{Session 1:} During the first session, we presented the study to participants and collected the age and gender data. We also administered the questionnaires concerning the curiosity trait, the perception of curiosity and device use frequency. We then run the reading abilities test and the question-asking fluency test (QA fluency test), see more details of the tests in the section \ref{section:measures}. We also took time to explain what divergent questioning means and highlighted the difference between divergent and convergent questions. Participants then went to the evaluation space of "Kids Ask" and began the interaction with a general domain-knowledge quiz relating to six themes. For each question, children had the choice to either skip the item, by clicking on the 'I don't know, I want to skip this question' button, or answer it. If they do choose to answer, they are asked to report their confidence level in their answer: they had a 5-Likert scale from 'Super not confident' to 'Super confident' (see Figure \ref{fig:Quiz}).

Once finished, participants chose their favourite topic among the ones they had during the quiz. They were told to choose carefully as this is the topic they will be working on during the next task.

\textbf{Session 2:} During the QA training session, participants were asked to read short texts relating to the theme they chose. They then started working on generating divergent-thinking questions relating to these texts using the cues provided by the agent. They also had audio players for each text to help them if they had reading difficulties. The texts were selected from online resources and children magazines (Sciences et vie Junior and Quelle Histoire) and were edited in order for them all to have six sentences and an average number of 109 words per text.

\begin{figure}
  \centering
  \fbox{
  \includegraphics[width=.9\linewidth]{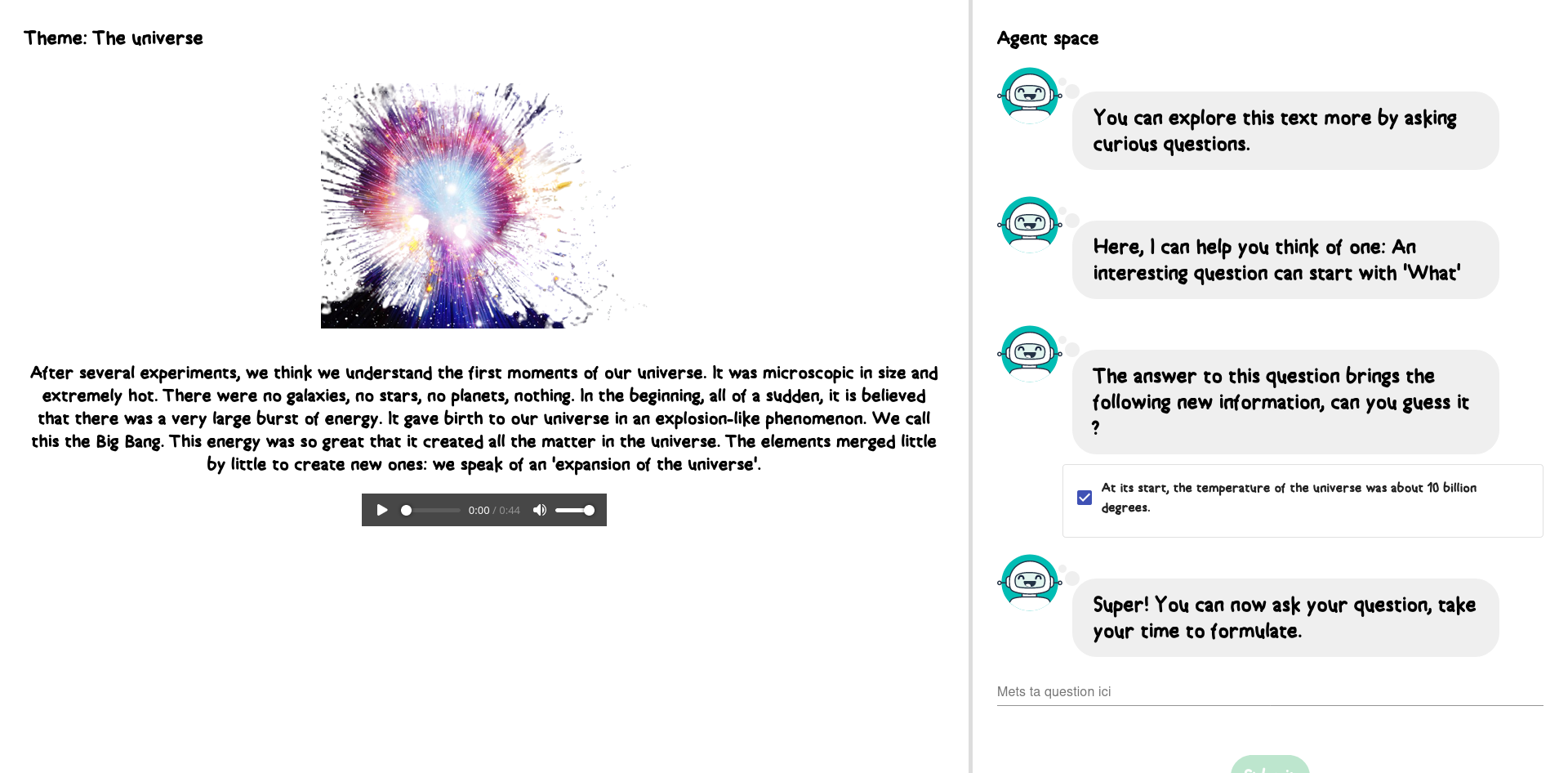}}
  \caption{\textbf{Agent behavior for the incentive conditions}}
  \label{fig:qa-c}
\end{figure}

\begin{figure}
  \centering
  \captionsetup{justification=centering,margin=1cm}
  \fbox{
  \includegraphics[width=.9\linewidth]{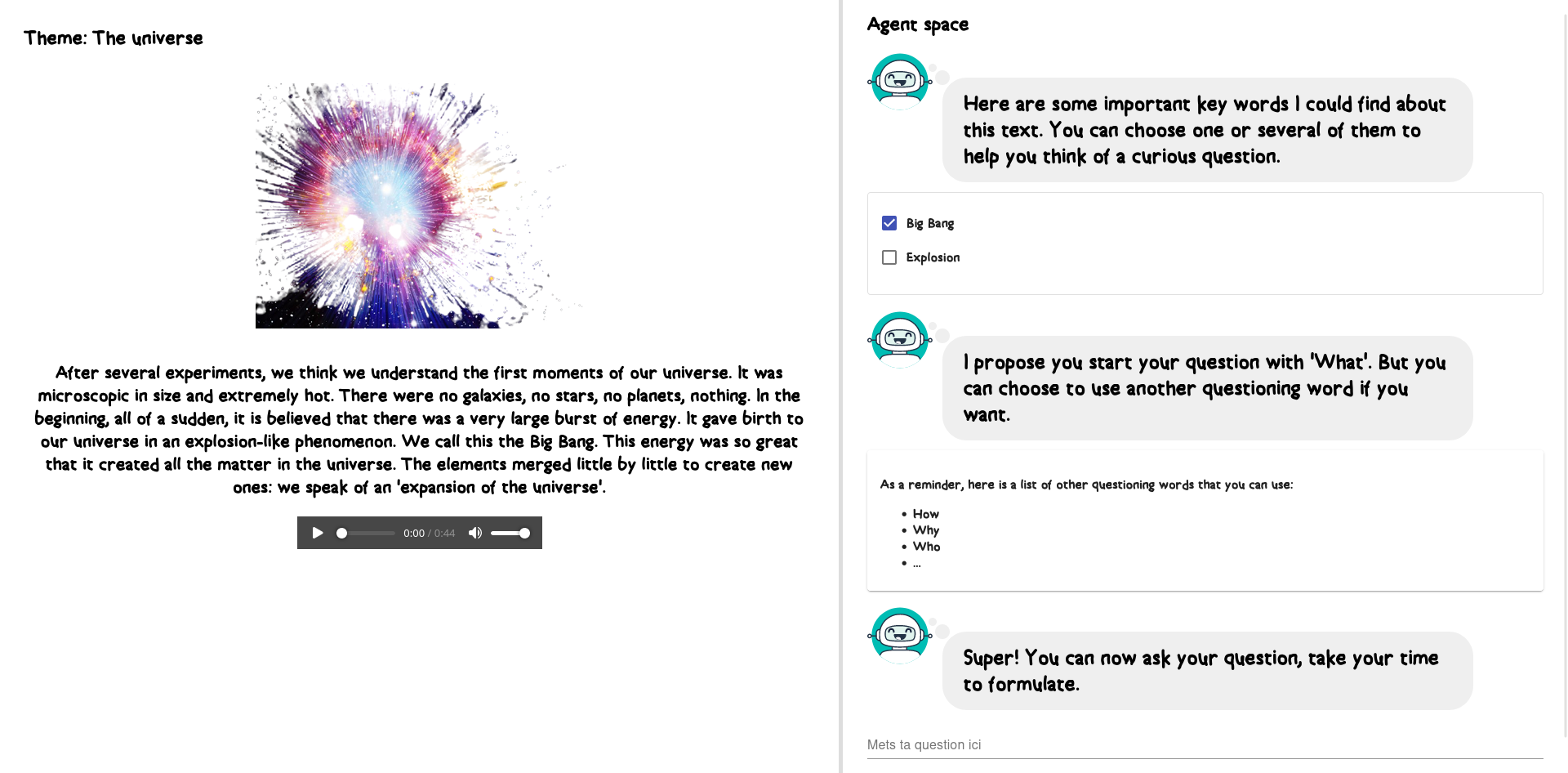}}
  \caption{\textbf{Agent behavior for the "open" condition}}
  \label{fig:qa-e}
\end{figure}

To begin working on a text, participants were asked to read or listen to it and then click on the ‘I finished reading’ button once they understand it. This button enables the ‘discussion’ with the agent, in the agent's space on the right section of the screen. As explained above, the agent helps the child generate divergent-thinking questions about the text by giving specific linguistic and semantic cues. The two incentive agents (hand-crafted and GPT-3-driven) had the same behavior; their only difference lied in the generation method of the cues presented. While the "open" had slightly different utterances to go with the type of cues it proposes. See Figures \ref{fig:qa-c} and \ref{fig:qa-e} for the difference between the conditions. 

During this session, children had to process three texts and generate six questions per text with the agent, making a total of 18 questions. They were not restricted in terms of time, apart from the session length. They had the application running on tablets and worked individually.


\textbf{Session 3:} Participants answered the post-intervention surveys for the general motivation, types of motivation and the task load. They also re-took the QA fluency test and the curiosity perception questionnaire they had pre-intervention (see measures timeline in \ref{fig:timeline} and further descriptions in the "Data collection and instruments" section \ref{section:measures}).

\section{Data collection and measures}
\label{section:measures}
In this section, we describe the different measures we use during this study : we start by the metrics used to evaluate GPT-3's output and then move to describe the participants' profiles and measures used to assess their divergent QA performance when having the training. Finally, we describe the learning experience measures and the training's pre-post effect.

\subsection{Evaluating of the cues generated by GPT-3}
\label{section:cues-measures}
In evaluating the cues generated by GPT-3, we use human annotations performed by two of the researchers that conducted this study, using specific grids for the linguistic and semantic cues as we describe below. It is to be noted that one of these annotators was an expert in using the evaluation grids whereas the second was a complete novice.

\subsubsection{The linguistic cues}
In studying the quality of the linguistic cues generated (i.e. the questioning words), we inspect two dimensions and compare them to what was generated for the hand-crafted agent : the variety of the words proposed and their complexity level. The variety of the question words proposed is an important observation for us in order to ensure that children do not receive the same words to use during the whole training. The complexity level is an interesting dimension to evaluate as studies such as~\cite{Humphries2015} show that compound questioning words (e.g. "What difference", "What if" ..) are more likely to lead towards divergent questions than simple words like "What", "Where" etc. So we compare the proportions of occurrence of this type of questioning words when generated by hand vs. by GPT-3.

\subsubsection{The semantic cues}
In scoring the quality of the cues generated by human experts or GPT-3, we evaluate their semantic relatedness to the text in question. This is an important measure given that the goal is to lead children to ask divergent questions that are closely related to the educational resource. For this we use a 5-scale Likert and run a human annotation; the items the grid we used can be found in the Appendix\ref{appendic:semantic-relatedness}.

We also take into account the divergence level of the cues, compared to the text's context. This will help us understand better children's divergent QA performance later on, because it helps understand if the support we're giving them is pushing towards divergent questioning in the first place. We also use human annotation to assess this aspect, following the grid described in~\ref{appendic:div-level}. 

\subsection{Evaluating children's QA performance}
Same as for the evaluation of the cues, we also perform human annotations to study children's questions, using new grids. The same coders annotated the questions and once again, one was an expert of using the grids and one was not.

\subsubsection{Participants' profiles}
We recruited 75 4th grade students from three classes in two French primary schools. They were between 9 and 10.5 years old and assigned either to the group 1 with the incentive hand-crafted agent (24 with 13 boys and 11 girls), group 2 with the incentive GPT-3-driven agent (26 with 13 boys and 13 girls) or to group 3 with the open GPT-3 agent (25 with 11 boys and 14 girls).

The groups were assigned with a pseudo-randomized method after collecting profile data regarding the age, digital device use frequency, curiosity related measures, reading fluency and the domain-knowledge scores (see \ref{section:measures} for more details of the measures). As shown in Table \ref{Tab:participants}, we had three balanced groups that were not different in terms of these initial profile measures, with the exception of the familiarity with using digital tools measure. But since the study was run in a controlled environment where researchers could help participants when they have any technical difficulties, we hypothesize that this difference does not have a significant effect of our measure of interest, i.e. capacity to formulate divergent questions. To ensure this, we run an ANCOVA analysis in order to investigate the effect of the participants' exposure to digital tools score on their divergent QA performance, between the three groups. Results showed that the interaction was the same between the three groups (p-value=0.23) suggesting that the exposure variable did not affect the equilibrium between our participants.

\begin{table}[h]
\begin{center}
\caption{Profile measures for the three experimental conditions}\label{Tab:participants}%
\begin{tabular}{@{}rcllll@{}}
\toprule
Measure&Group 1&Group 2&Group 3&p-values\\
\midrule
Age & 9.4 $\pm $ 0.43& 9.56 $\pm $ 0.38 & 9.52 $\pm $ 0.44 & 0.31\\
Device use frequency & 27 $\pm $ 7.66 & 31.39 $\pm $ 6.12 & 32.57 $\pm $ 5.28 & 0.008* \\
Curiosity trait & 27.2 $\pm $ 4.38 & 29.69 $\pm $ 4.53 & 28.36 $\pm $ 4.75 & 0.16\\
Perception of curiosity & 37.54 $\pm $ 6.72 & 38.3 $\pm $ 6.66 & 33.92 $\pm $ 9.12 & 0.09 \\
Reading ability & 279.55 $\pm $ 93.02 & 297.37 $\pm $ 104.61 & 293.19 $\pm $ 73.6 & 0.77 \\
Ability to ask div. questions & 1.33 $\pm $ 0.8 & 2 $\pm $ 1.17 & 2.2 $\pm $ 0.93 & 0.11 \\
Domain-knowledge quiz score & 6.9 $\pm $ 4.6 & 8.77 $\pm $3.58 & 8.72 $\pm $ 4.05 & 0.23\\

\end{tabular}
\end{center}
\end{table}

\subsubsection{Percentage of divergent questions generated}
\label{section:percentage-questions}
This measure consists of counting the total number of the correct questions generated during the training and computing the percentage of the divergent questions amongst them.  If a question is repeated, it is only taken into account once. Questions that did not use the agents' cues were taken into account if they were still relevant to the general context of the text under discussion. See more details about our acceptance criteria in the appendix~\ref{appendic:qa-evaluation}.

Once a question is accepted, we evaluate whether or not it is divergent by checking if its answer is explicitly stated in the text. This is based on Gallagher's classification ~\cite{Aschner1963}. \textbf{Example:} The question “What was the temperature of the universe at its beginning?” is considered to be divergent, whereas "What caused the birth of the universe?" is considered convergent as the answer to it is explicitly stated in the text.

All data was anonymized: the coders could only see the identifiers that children were given randomly at the beginning of the intervention. The inter-rater reliability was of 87.6\%.

\subsubsection{Quality of the questions generated}
\label{section:scores-questions}
As mentioned in the beginning of this article, we find several taxonomies to classify questions and as many ways to annotate them. We chose to adopt one of the most used ones by Wilen et al~\cite{wilen1991} that differentiates 2 main types of questions (convergent and divergent) at two levels (low and high) according to semantic and syntactic criteria related to the expected answer and/or the question formulated. In this distinction, it is suggested that syntactic and semantic dimensions are tightly related : divergent questions are associated with more complex syntactic constructions. This classification is inspired from Gallagher's quotation grid in~\cite{Aschner1963} but adds the syntactic dimension in order to have a qualitative aspect of the question. The grid scores a question from 2 to 8 points (see details in appendix~\ref{appendic:syntax-scores}).

For each participant, we calculated an average score for all the questions generated during the training. The inter-rater reliability of 76.8\% and for data points where we saw difference between the too coders, we used an average score.

\subsection{Evaluating children's learning experience}
\label{section:le-measure}
\subsubsection{Motivation measures}
We use Vallerand's scale to assess the intrinsic and extrinsic motivational mechanisms~\cite{vallerand1989}. The scale is composed of three sub-scales that differentiate: intrinsic motivation (possible scores from 0 to 9 points), extrinsic motivation (possible scores from 0 to 9 points) and amotivation (possible scores from 0 to 3 points). All questionnaire items are yes or no questions.

\subsubsection{Cognitive load}
The workload of the task, as perceived by the participants was assessed using the NASA-TLX workload multi-dimensional scale developed in~\cite{Nasa-tlx}. Information about the intensity of six workload-related factors are used in order to estimate a reliable measure of workload: mental demand, physical demand, temporal demand, performance, effort and frustration. 

\subsection{Evaluating the training's effects}
\subsubsection{On children's divergent QA skills}
\label{section:qa-fluency}
In measuring the short-term effect of our training on children's ability to ask divergent questions in different contexts, we use an offline test inspired from creativity studies such as in~\cite{diakidoy2002}. This test consists of giving participants a short text to read then asking them to write as many questions about it as they can, within 2 minutes.

Using the same annotation we described in~\ref{section:percentage-questions}, we evaluate the questions children generated during this test and investigate whether they tended to ask more divergent questions after the "Kids Ask" training or not, and whether this change depended on their experimental condition.

\subsubsection{On children's perception of curiosity}
\label{section:ciac}
In order to assess if and how children's attitude towards curiosity in general and asking questions in particular has evolved with our training, we use Post's validated CIAC questionnaire~\cite{Post2015}. Indeed, this is an important measure for us as we think that one of the relevant brakes that can keep children from asking questions in the classroom is their negative perception of this behavior.

We also use one of this questionnaire's sub-scales to evaluate how children's perception of their own competency in generating high-level and curious questions has evolved after the training. This is also an important measure as following theories such as in~\cite{oudeyer2008}, we suppose that observing the learning progress on a certain task can reinforce the individual's intrinsic motivation to engage in it further more.

\section{Research goals and hypotheses}
This work studies the implementation of GPT-3-driven CA that aims to help children generate divergent questions by proposing specific semantic and linguistic cues. To do so, we test two different approaches, an incentive one and an open one. We evaluate their efficiency using human annotations of the generated output and children's performance when using this training. We hypothesize that:
\begin{itemize}
\item GPT-3 will be able to generate correct and efficient curiosity-eliciting cues, compared to hand-crafted cues.
\item The GPT-3-driven incentive agent will lead children to similar divergent question-asking performances as the hand-crafted incentive one.
\item The open agent, compared to the two incentive agents, will lead children to ask more divergent questions.
\item The divergent QA performance will be a strong positive predictor for children's curiosity traits, as reported by their parents.
\item The divergent QA training, irrespective of the agent's implementation, will enhance children's perception of curiosity. 
\item The divergent QA training, irrespective of the agent's implementation, will enhance children's ability to ask divergent questions.
\end{itemize}

\section{Results}
\subsection{How did human experts evaluate GPT-3's output ?}
In this section, we evaluate GPT-3's performance in generating curiosity-eliciting cues using human annotations. We start with general observations then move to investigating the quality of the linguistic cues then the semantic ones.

\subsubsection{How did human experts evaluate the general relevance of the cues generated by GPT-3?}
\label{section:general-res}
As a first step, we start by verifying GPT-3's outputs with regards to offensiveness and general relatedness to the context of the educational resource in hand. For the offensiveness measure we run a human annotation with a 5-Likert scale (from "Not at all offensive for a 10-year-old child" to "very offensive for a 10-year-old child"). Results showed that 100\% of the data generated was scored as no at all offensive (m=5 ; SD =0).

We then evaluated the output generated depending on its relevance to the general context of the text (e.g. for a text about the Big Bang, data-points about the universe in general are considered relevant, however output about different topics like the sports ... are considered not relevant). Here again, we run a human annotation and saw that 100\% of our generated data was annotated as relevant to the task.

\subsubsection{How did human experts evaluate the quality of the linguistic cues generated by GPT-3 ?}
We evaluate GPT-3's ability to generate appropriate and high-level questioning words by comparing its output to what was generated by human experts. We inspect two dimensions : the variety of the words proposed and their complexity level.

\begin{figure}
\centering
\captionsetup{justification=centering,margin=1cm}
\includegraphics[width=1\columnwidth]{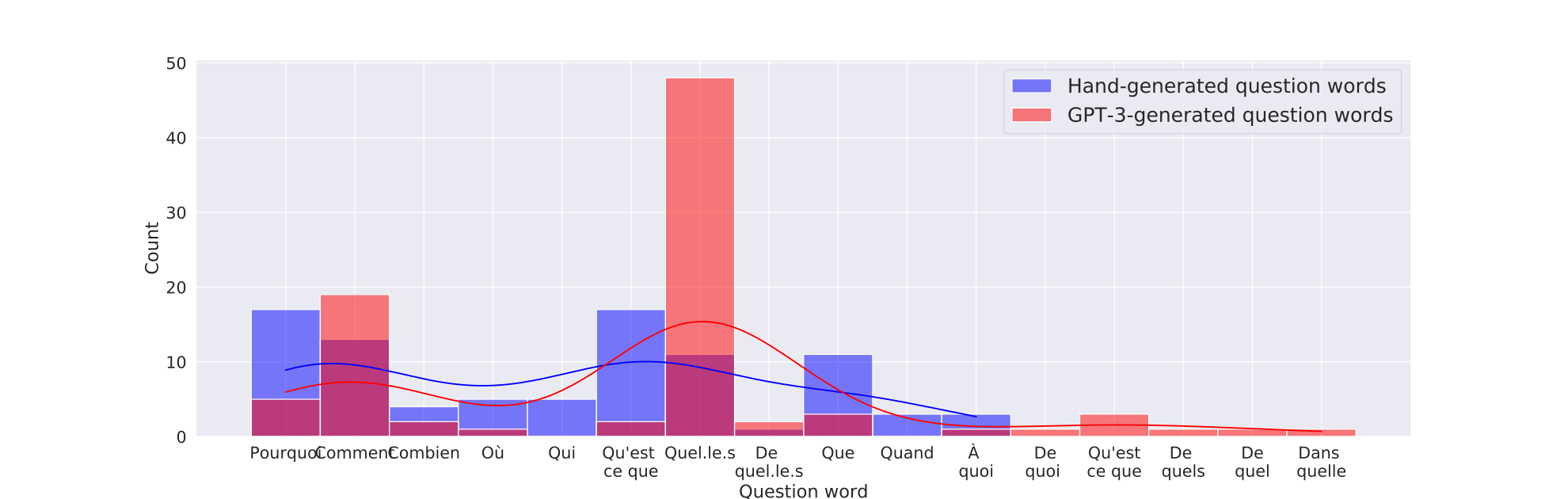}
\caption{\textbf{Generating the linguistic cues : Distribution of the question words proposed in the "Kids Ask" platform.}}
  \label{fig:cues-eval-qw}
\end{figure}

To assess this, we plot the histogram of the cues generated by hand and by GPT-3 and investigate the similarity between the variances of the two distributions (See Figure \ref{fig:cues-eval-qw}). We use a Levene's test given that the question words generated by hand deviated from normality. Results showed no support to reject similarity between the two conditions (p-value= 0.65). 

We then studied the complexity levels of these cues by comparing the proportions of occurrence of compound words when generated by hand vs. by GPT-3. As mentioned above, this is an important aspect as compound questioning words are shown to lead to more divergent questions. The 2-sample z-test showed no support to reject the hypothesis of having two similar proportions (p-value = 0.51).

With these results, we see similar behaviors between human experts and GPT-3 in generating appropriate questioning words that can lead to divergent QA.

\subsubsection{How did human experts evaluate the quality of the semantic cues generated by GPT-3 ?}
\begin{figure}%
    \centering
    \subfloat[\centering \textbf{The semantic relatedness of the cues generated by GPT-3 is similar to what was produced by hand.}]{{ \includegraphics[width=5.5cm]{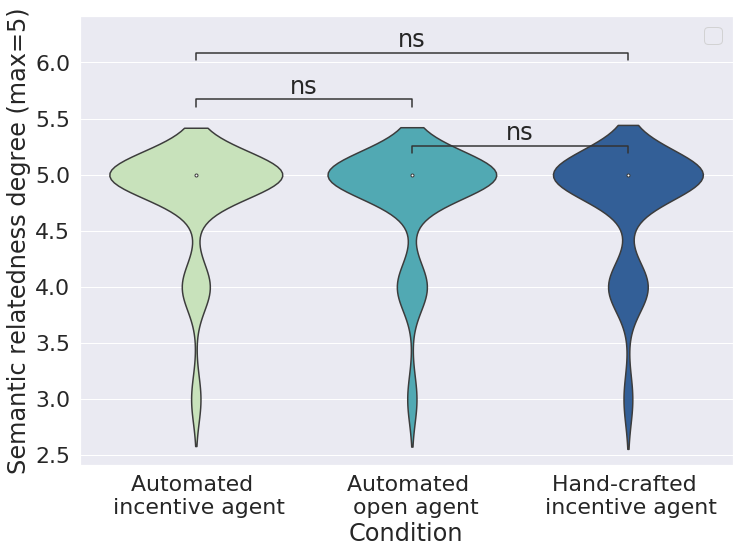} }}%
    \subfloat[\centering \textbf{The divergence level for the incentive semantic cues was similar for the hand-crafted and the GPT-3-driven versions.} ]{{\includegraphics[width=6.5cm]{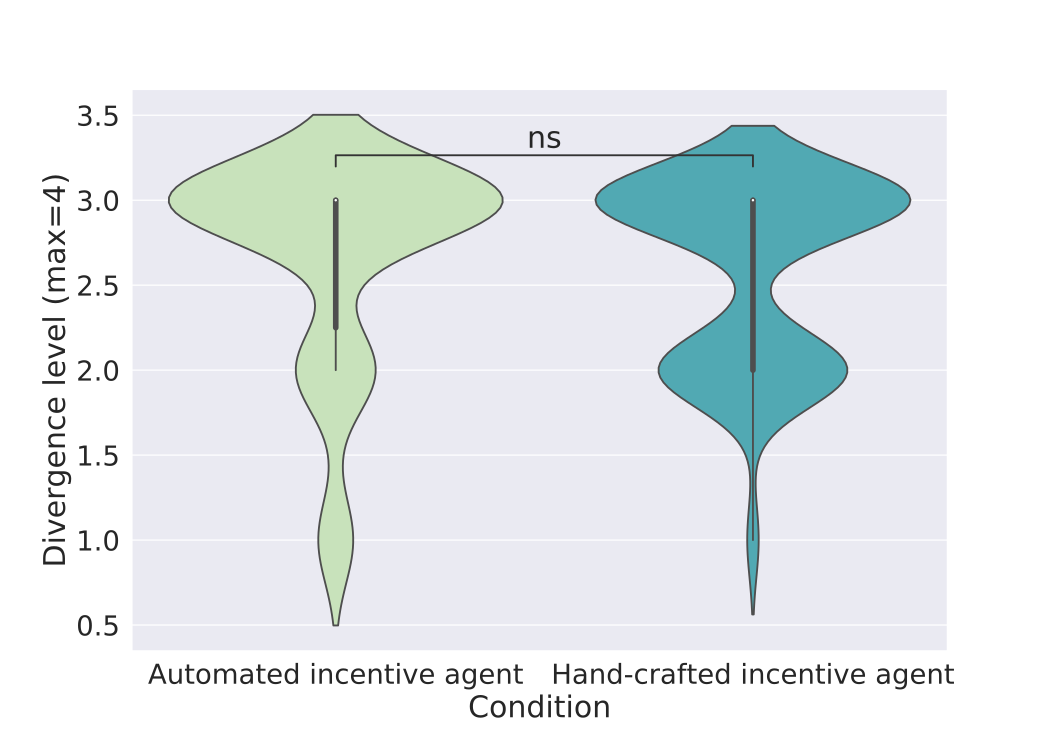} }}%
    \caption{Generating the semantic cues : Evaluation with respect to the semantic relatedness to the context and the divergence level.}%
    \label{fig:cues-eval-sc}%
\end{figure}

Here again, we base our evaluation on comparing specific dimensions of the semantic cues generated by GPT-3 for the incentive and the open agents to those generated by hand. We start by investigating the semantic relatedness degree of the cue to the context of the proposed educational resources, as described in the Measures Section\ref{section:cues-measures}. As seen Figure\ref{fig:cues-eval-sc}, we see no significant difference between the three conditions for this measure. The one-way ANOVA test failed to exclude similarities between at least two groups (F(2,74)=0.41, p-value = 0.66), with a power test of 0.98. Assumptions about the normality of our distributions, homogeneity of variances and independence between the observations were confirmed pre-running the test.

We also study the divergence levels of the proposed cues for the two incentive agents. It makes sense to assess this aspect given that the incentive agents propose specific cues that will heavily condition children's questions. Thus, if the proposed cues are already mentioned in the text i.e., convergent, they will automatically lead children to generate convergent questions. However, this is not the case for the open agent's cues, i.e., the keywords, as the questions' divergence levels do not rely on the words themselves but on how children choose to use them. For this reason, we only assess the cues' divergence levels for the two incentive agents, by running a human annotation following the grid described in~\ref{appendic:div-level}. As it can be seen in Figure\ref{fig:cues-eval-sc}, we run a t-test and saw the two distributions did not differ (t=0.9, p-value=0.37 and power=0.14).

Similarly to the linguistic cues, here we also see results suggesting that our approaches were successful in generating divergent semantic cues that are closely related to the context in question, compared to what we had in the hand-crafted condition.

\subsection{How did human experts evaluate children's QA performance during the training ?}
Moving on to studying the children's interaction with the system, we start by looking into their QA performance during the training, within the different experimental conditions. We then investigate the relationship of this performance with the curiosity trait scores and evaluate the learning experience in terms of its pleasantness and its perceived cognitive load. We finally report the mid-term effect of our training on children's ability to generate divergent questions in a different context as well as their perception of their own competency on this skill.


\subsubsection{How did children's divergent QA skills vary across conditions ?}
\label{section:children-div-score}
In a first step, we start by evaluating the general usefulness and accessibility of the cues proposed by the CAs by seeing how often children actually chose to use them to generate their questions. Results showed no difference between the two incentive agents for the percentage of questions where they used the agents' cues (M=77.28\% and SD=23.08 for the hand-crafted agent; M=76.12\% and SD=23.88 for the automated agent with a p-value for the T-test of 0.86 and a power test of 0.05). However, children with the open automated condition used the agent's cues more often (M=91.78\%, SD=8.83) and were significantly different from the hand-crafted incentive agent (p-value= 0.03, power=0.85) and the incentive GPT-3-driven one (p-value=0.005 and power=0.81); suggesting that the keywords-based cueing strategy may have been more accessible for children.

\begin{figure}%
    \centering
    \subfloat[\centering \textbf{Participants with the incentive agents (hand-crafted or GPT-3-driven) had similar divergent QA performances. Those who had the automated open agent had a significantly better performance than the two incentive groups.}]{{ \includegraphics[width=5.5cm]{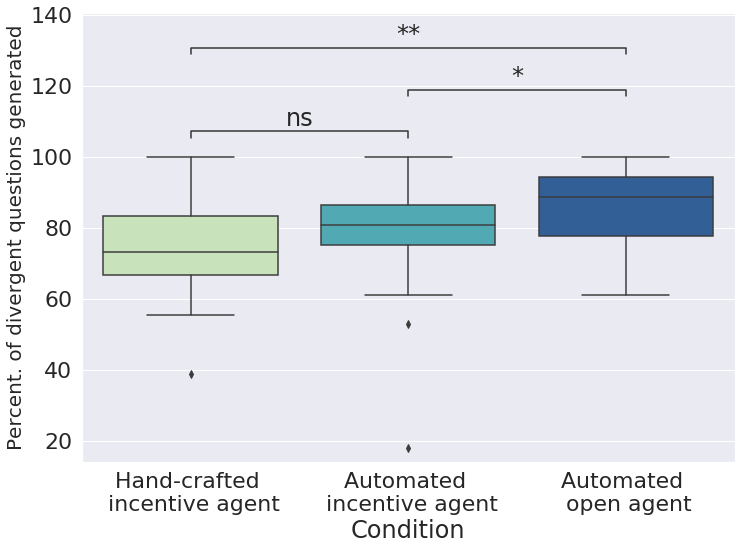} }}%
    \subfloat[\centering \textbf{The average quality score per question generated was similar among participants from the three different conditions.} ]{{\includegraphics[width=5.5cm]{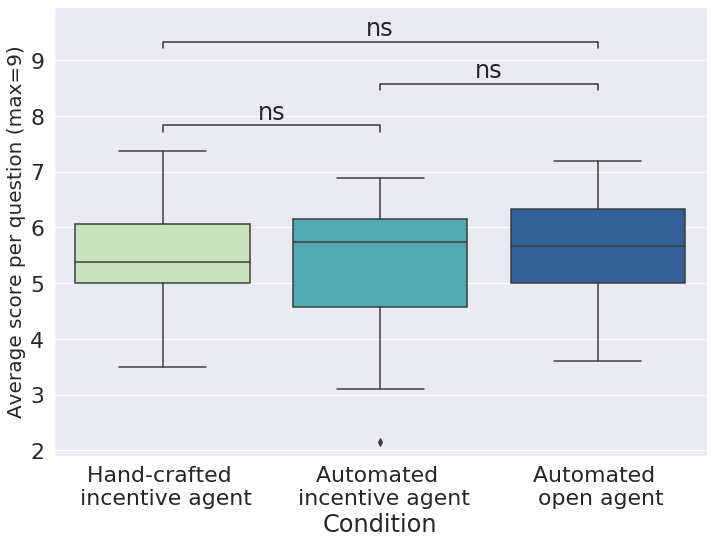} }}%
    \caption{Question-Asking performances during the training, using the QA-training workspace}%
    \label{fig:qa}%
\end{figure}

Moving on, we analyze children's QA performances by comparing the percentage of divergent questions generated between the three experimental conditions. These percentages are computed as explained in Section \ref{section:percentage-questions}. We start by performing a one-way ANOVA test between the three groups; results show indeed a significant difference in the performances (F(2,72)=4.11, p-value=0.02). We then perform T tests in pairs : we see no significant difference between the two incentive agents, i.e. the hand-crafted one and the GPT-3-driven one (t=0.88, p-value=0.38 and power=0.14). However, the group that interacted with the automated open agent had a significantly better performance than the Hand-crafted incentive agent (t=-3.17, p-value=0.003 and power=0.82) and the GPT-3-driven incentive agent (t=-1.88, p=0.04 and power=0.45); see Sub-Figure (a) of Figure \ref{fig:qa}. The normality and the homogeneity of variances of our data were verified pre-running our tests.


\subsubsection{How did children's questions quality vary across conditions ?}
\label{section:children-qa-score}
We also investigated the quality of the questions generated, following the standardized grid described in \ref{section:scores-questions}. These scores give us an idea about the syntactic and semantic quality of the children's questions. Results showed no significant difference between the two incentive agents (t=0.57, p-value=0.57 and power=0.08). The same observation was also found between two automated agents, i.e. incentive vs open (t=-0.81, p-value=0.41 and power=0.12) and between the hand-crafted incentive agent and the the automated open one: t=-0.29, p-value=0.77 and power=0.06. See Sub-Figure (b) of Figure\ref{fig:qa}.

\textbf{First conclusion:} Results from the two previous subsections (\ref{section:children-div-score} and \ref{section:children-qa-score}) show that children who interacted with incentive hand-crafted CA had an overall similar divergent QA performance to the GPT-3-driven one suggesting the validity of the GPT-3 cueing approach we used to generate the 'closed' cues. Furthermore, they suggest that using GPT-3 to generated more 'open' cues, i.e. keywords, may be a stronger strategy to help stimulate children's divergent questioning. Indeed, this interaction helped children formulate more divergent questions while maintaining similar syntactic scores even though they had no semantic formulations for their cues.

\subsubsection{How did children's curiosity trait correlate with their divergent QA ?}
In this subsection, we study the interaction between children's divergent QA and their curiosity trait and how it is affected with the interactions with the different CAs. The results show a significant correlation between the two only for the automated open agent. Details about the statistical tests can be found in the appendix~\ref{appendix:cur-div-qa}.

These results may posit that our open agent is more associated with curious thinking given that it leaves more space to children to formulate questions of their own choice and thus, to express their own curiosity. However, in the case of the two incentive agents, we saw no significant correlations between their performance in the task and their curiosity trait. This can be explained by the idea that, given that children were restrained to think of pre-defined specific questions, their task was more of a social curiosity training rather than an idiosyncratically one, as explained in Section\ref{section:choice-cues}. 

\subsection{How did children perceive their learning experience with "Kids Ask" ?}
\subsubsection{How motivated were children to use the platform ?}
\begin{figure}%
    \centering
    \subfloat[\centering \textbf{Participants for the the three groups showed significantly more intrinsic motivation than amotivation or extrinsic motivation.}]{{ \includegraphics[width=5.5cm]{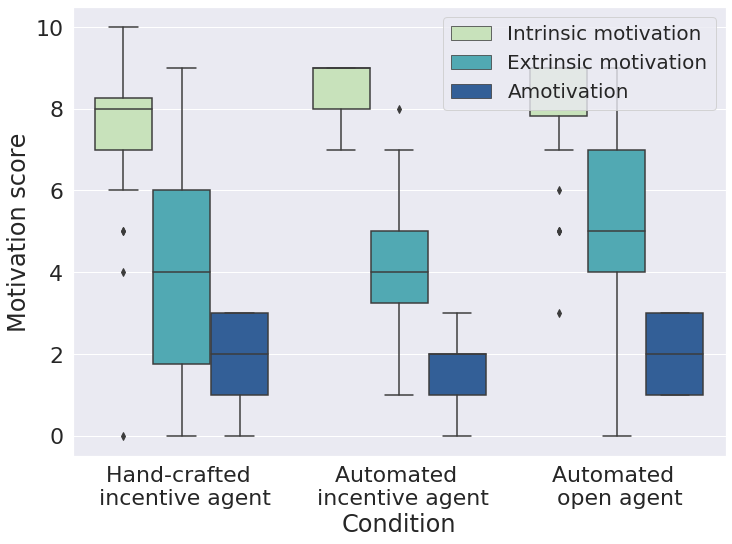} }}%
    \subfloat[\centering \textbf{The cognitive load of the task was similar for the three different conditions.} ]{{\includegraphics[width=5.5cm]{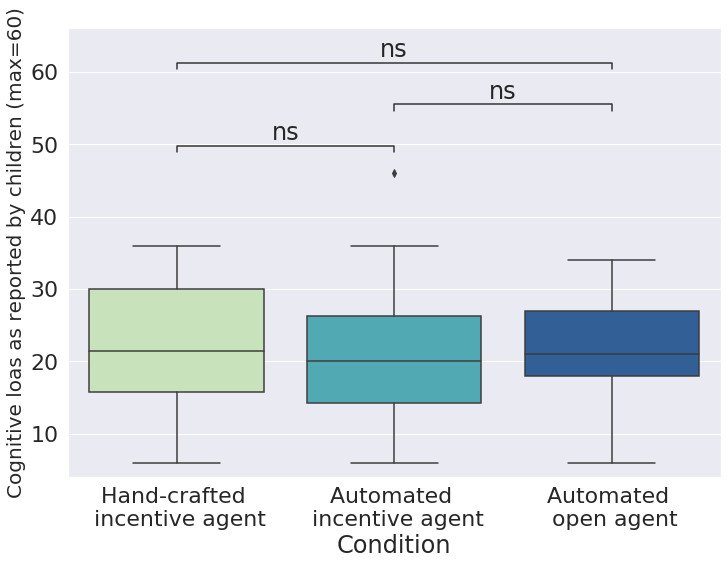} }}%
    \caption{Learner experience measures}%
    \label{fig:le}%
\end{figure}

We see no difference between the children's intrinsic motivation during the training with the three different agents. However, we saw that the three groups were significantly more intrinsically than extrinsically motivated to do the task (t=3.3, p$<$0.001, power=0.98 for the hand-crafted incentive agent, t=12.02, p$<$0.001, power=1 for the automated incentive agent and t=3.9, p$<$0.001, power=0.97 for the automated open agent); see sub-figure (a) in Figure\ref{fig:le}. 

\subsubsection{How did children perceive the cognitive load of the task ?}
On a final note, we also investigated the cognitive load as perceived by children, following the Nasa-tlx scale as described in Section~\ref{section:le-measure}. The aim here is twofold: 1) evaluate whether or not children perceived that the answers provided by GPT-3 were harder to assess than those provided by humans. 2) see if the lack of a semantic formulation in the open cues for the open automated agent makes the question-generation task harder for children. To address these goals, we perform pairwise t-tests. Results showed no significant differences between the two incentive agents (t=-0.4, p=0.1, power=0.06) or between the incentive and open agent (t=-0.3, p-value=0.7, power=0.06). These results suggest indeed the accessibility of GPT-3's output even when no semantic formulation was given (i.e. only keywords).

\subsection{How did the "Kids Ask" training affect children's curiosity ?}
\subsubsection{How did children's ability to ask divergent questions change after the "Kids Ask" training ?} 
\label{section:pre-post-qa}
We were interested in studying whether there was a short-term effect of our intervention, i.e. if the training helped children be better at asking divergent questions when they are put in another context than "Kids Ask". For this reason, we assess children's ability to ask divergent questions before and after the intervention using the offline test described in section \ref{section:qa-fluency} and compare the three different conditions.

\begin{figure}%
    \centering
    \subfloat[\centering \textbf{Participants from the three conditions were able to improve their divergent QA abilities after the ”Kids Ask” interaction, as shown by the divergent QA
fluency test pre- and post-training.}]{{ \includegraphics[width=5.5cm]{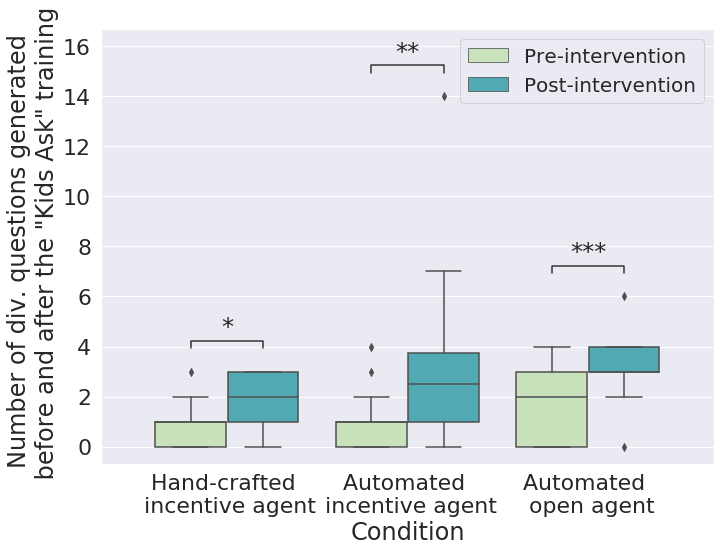} }}%
    \subfloat[\centering \textbf{Children’s perception of their QA self-efficacy changed more positively with the intervention for those who interacted with the automated agents.} ]{{\includegraphics[width=5.5cm]{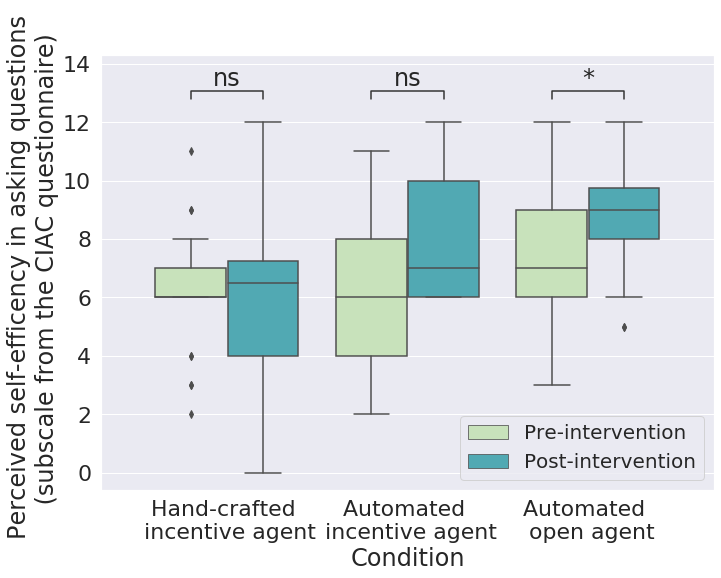} }}%
    \caption{Intervention effect : pre- and post-training measures}%
    \label{fig:pre-post}%
\end{figure}

We performed a two-way repeated measures ANOVA test in order to investigate the impact of the time on the divergent QA performance between the three conditions. Results showed a statistically significant effect of the time (F(2,72)=46.95, p-value$<$0.001) and the condition (F(2,72)=3.9, p-value=0.002), with no effect of the interaction condition:time on the performances (F(2,72)=2.23, p=0.11); see Sub-figure (a) in Figure\ref{fig:pre-post}. Hypotheses concerning the absence of outliers and normality of the divergent QA-fluency scores were verified pre-running the test.

These results show that participants from the different groups benefited from the intervention, suggesting the general validity of the cueing approaches used and the benefits of using CAs to train divergent children's divergent QA skills in general. 

\subsubsection{How did children's perception of curiosity change after the "Kids Ask" training ?} 
\label{section:pre-post-ciac}
In analyzing the effect of our intervention on children's general attitude towards epistemic curiosity, we run a two-way mixed ANOVA test to investigate the difference in the CIAC scores(Section \ref{section:ciac}) before and after our intervention, between the different conditions. Results show a non-significant interaction between time and the condition type (F(2,72)=3.87 , p-value=0.06) but reveal a significant change in the scores for all participants pre- and post-intervention (time Effect: F(1,72)=15.74, p-value$<$0.0001). This suggests the general efficiency of our proposed method in enhancing children's perception of curiosity in the classroom.

We then investigated the Self-Efficiency sub-scale of the CIAC questionnaire which measures how children perceive their own skills in asking relevant and complex questions. As shown the sub-figure (b) in Figure~\ref{fig:pre-post}, the three groups had better scores post-intervention and the interaction between time and the groups was significant (F('2,72)=0.038, p-value=0.04). The post-hoc pairwise tests showed a significant increase in this score only for the group that interacted with the open GPT-3 driven agent (p=0.014). This result aligns well with what's discussed in the design rationale section~\ref{section:choice-cues} where we suggest that giving children tasks with a larger margin of choice and autonomy can effect their perception of their self-efficiency. 

\section{Discussion}
With this work, we investigate using prompt-based learning with the pre-trained LLM GPT-3 to generate the pedagogical content for a curiosity training that aims to develop divergent QA skills in children. We investigate the validity of our approach by using human annotations to score this content and compare it to what was generated by hand. 

The proposed prompt-based method showed positive results for GPT-3's ability to generate relevant curiosity-eliciting cues, both in terms of linguistic quality, semantic relevance and pedagogical value for children. More specifically, our LLM-content-based curiosity training showed more efficiency in eliciting children's divergent QA when we had open cues leading to several possible questions, rather than the incentive ones that led to one specific and pre-defined question. These observations were somehow expected given that the open agent was designed with the intention to propose a student-directed training. Indeed, we hypothesize that such a setting will lead children to work on questions that correspond to their competency level which will result in better perception of the self-efficiency in the task ~\cite{METCALFE202040} and therefore, in more engagement and learning progress~\cite{oudeyer2008, Ten2021}. 

In a further step, and in an attempt to reinforce previous findings that link curiosity trait to divergent QA, we investigated the relationship between children's performance during our divergent QA training and their curiosity trait as reported by their parents. Our results showed a strong correlation only for the group that had the open cues. Several ideas can be posited to explain these findings : first, it is possible that children who are more curious by trait, since they are more familiar with asking questions without any syntactic support, found in the open agent a more favorable condition to express their curiosity. Second, it is possible that giving children QA activities that they can transpose to their ZPDs helped their curiosity. We lean more towards the second explanation since we saw a significant increase in children’s perception of their QA skills in the open conditions. This is an interesting finding as it addresses the question of proposing new behavioral measures for curiosity that are suited to e-learning environments, rather than using the classic self-report measures.

Finally, this current implementation also addressed some of the challenges we raised earlier that are shown to keep children from asking questions in the classroom (i.e. difficulty to find the syntactic formulation for divergent questions, fear of negative judgment and lack of metacognitive efficiency). Indeed, and as we explain this in the Design rationale section~\ref{section:choice-cues}, we addressed the first challenge by choosing to provide linguistic cues along the semantic ones in order to help children practice the syntactic formulation of divergent questions with the help of the CA. Consequently, and as shown in the results sub-section~\ref{section:pre-post-qa}, our QA fluency test showed that children were indeed able to ask significantly more syntactically-correct divergent questions in the post-intervention test, for all three experimental conditions. Suggesting, thus, the general validity of the training method. Moreover, this work also tempted to address the negative perception of curiosity challenge by proposing a novel and attractive tool to practice this skill (interaction with a CA, use of digital devices, etc). In fact, we saw high motivation scores to use the tool for the three conditions with a predominance of the intrinsic component. Indeed, our training effect showed to be positively significant on the CIAC measure pre- and post-intervention for all conditions (sub-section~\ref{section:pre-post-ciac}). This measure captures children's image of curiosity, including their fear of their classmates' negative judgment when asking questions in the classroom.

These results suggest the efficiency of our general curiosity-eliciting method in addressing some of the brakes that can keep children from expressing their curiosity, independently from the experimental condition.

\section{Limitations and future directions}
One drawback to our current implementations is the lack of feedback we give children about their performance during the task. Indeed, and even though we see enhanced perceptions of the self-efficiency, we do not know if this is directly connected to their perception of their performance or to other factors, like the repeatability of the exercise, bias in the self-report measure, etc. Therefore, one future direction for this work will be to explore ways to use LLMs in general in order to analyze children’s questions and give them real-time feedback about their relevance, their divergence level and their syntactic construction. Recent work such in~\cite{Xingdi2022} has explored using prompt-learning methods to evaluate the generation of convergent questions and showed rather encouraging results. Therefore, one possible track can be to take inspiration from their methods and explore adapting them to divergent questions. 

Another factor to be considered while evaluating this training is its short duration in time. Indeed, children were only asked to do this task during one short session which may have facilitated their engagement and played a role in the positive results we saw. Therefore, one future direction for this work is to propose a longer version of our training and investigate its long-term impact on children's curiosity-driven behaviors. It is also to be noted that, as mentioned above, the annotations of all the data generated during this training was performed by two of the researchers that led this study; it is therefore possible that some bias has been introduced in the scores, even though the inter-rater measures between the two were rather satisfying.

On another hand, even though prompt-based learning with GPT-3 is quite an easy method to implement, it does still require some knowledge about prompt-engineering in order to generate relevant output. Indeed, we hypothesize that we won't see the same results if we have used less precise and concise prompts and different settings. This leads us to think about the different trainings we should provide teachers with in order to ensure good practices and satisfying results.

Moreover, and as mentioned earlier in the "Ethical considerations" section, our implementation raises some educational challenges. Indeed, having children use LLM-based systems in general without knowledge about their best practices and without developing a critical thinking when interacting with them can be challenging. For example, it can reduce their creativity, engagement when resolving tasks, etc, making the system loose its potential benefits. In this direction, we are planning to develop pedagogical interventions in order to train children's intellectual humility and critical thinking when using LLMs.

Finally, this work did not explicitly address children's metacognitive efficiency. Indeed as mentioned earlier, this is an important skill that can modulate children's curiosity-driven behaviors and keep them from seeking new information. For this reason, we started working on the implementation of new pedagogical trainings that target specific metacognitive skills in order to evaluate their effect on developing curiosity.

\section{Conclusion}
This work is a part of a larger project attempting to develop technology-enhanced approaches to train curiosity in children. The study proposes to address some of the brakes that keep them from seeking new information autonomously such as linguistic barriers and negative opinions towards curiosity. More particularly, we carefully investigate the potential of LLMs to generate pedagogical content and propose a novel GPT-3-driven system that trains divergent QA using simple and efficient prompt-based methods.

Our results, including a field study, showed the benefits of using such tools to engage children in divergent QA tasks and improve their curiosity perception and QA fluency. This work motivates the implementation of such curiosity-eliciting approaches both in normal classroom settings and in e-learning environments using NLP methods.

\section*{Acknowledgements}
{This work has been funded by the educational technologies start-up EvidenceB
and the French National Association of Research and Technology (ANRT). The authors thank the teachers who participated in this study and the research team members who helped conduct the experiments in classes.}

\section*{Conflict of interest}
{The authors declare that the research was conducted in the absence of any
commercial or financial relationship that could be construed as a potential conflict of interest.}

\begin{appendices}
\section{Grid for the semantic relatedness measure}
\label{appendic:semantic-relatedness}
Likert scale used to annotate the semantic relatedness of our cues with respect to the text in question :
\begin{itemize}
    \item 1 point if the cue is not at all related to the text's context
    \item 2 points if the cue is not related to the text's context
    \item 3 points if the cue is somehow related to the text's context
    \item 4 points if the cue is related to the text's context
    \item 5 points if the cue is super related to the text's context
\end{itemize}

\section{Grid for the divergence level measure}
\label{appendic:div-level}
Likert scale used to annotate the divergence level of our cues with respect to the text in question :
\begin{itemize}
    \item 1 point if the cue is explicitly stated in the text
    \item 2 points if the cue is not explicitly stated but can be implied from the text
    \item 3 points if the cue is not at all stated in the text
\end{itemize}

\section{Criteria for accepting children's questions}
\label{appendic:qa-evaluation}
A question is considered correct and taken into account if:
\begin{itemize}
    \item It is a question --- and not a statement.
    \item It is related to the text being studied. 
    \item It is not repeated more than one time.
    \end{itemize}
Questions that did not use the agents' cues were still taken into account if they were still relevant to the general context of the text under discussion.

\textbf{Example:} For a text about the Big Bang, a linguistic cue ‘\textbf{What}’, a semantic incentive cue “\textbf{At its start, the universe's temperature was about 10 billion degrees}” or an open semantic cue "\textbf{Big Bang, explosion}", we accepted  questions such as: \begin{itemize}
    \item “What was the temperature of the universe at its beginning?” for the incentive agent or "What caused the Big Bang explosion" for the open agent: both questions use the agents propositions and are related to the text.
    \item “How does an explosion occur?” : used one of the agent's keywords and changed the questioning word but is still related to the text.
    \item "What does 'microscopic' mean?": For both agents, the question here does not use the cues proposed, but is highly relevant to the text.
\end{itemize}  
However, data such as the following was not accepted:
\begin{itemize}
    \item “It is the temperature when the universe began.“ : this is a statement and not a question.
    \item "What is a robot composed of?" : this is not related to the Big Bang text.
    \item "What are dinosaurs?" : this is not a serious attempt.
\end{itemize}. 

\section{Syntactic scores for evaluating children's questions}
\label{appendic:syntax-scores}
In assessing the questions syntactic quality, we use the grid inspired from Aschner's classification in~\cite{Aschner1963}. The grid calculates a question's score like the following:
\begin{itemize}
    \item One point if the question is high-level: the answer to this question is not a simple fact (example: 'How big is a dinosaur?') but requires to explain a mechanism, a relationship etc (example: 'Why were dinosaurs so big?').
    \item From 1 to 4 points, based on the syntactic construction of the question : \begin{itemize}
        \item 1 point for a ‘closed’ or declarative question (example: "Dinosaurs were big?").
        \item 2  points for questions with questioning words in the middle of the sentence (example: "The dinosaurs were how big?").
        \item 3 points for a question without an interrogative formulation (example: "Why the dinosaurs are big ?").
        \item 4 points for a questioning word in the beginning of the sentence that has interrogative syntax (example: "Why are dinosaurs big?").
    \end{itemize}  
    \item From 1 to 3 points, based on the use of questioning words : \begin{itemize}
        \item 1 point for a declarative question i.e., with no questioning word (example: "Dinosaurs were big?").
        \item 2 points for questions with ‘Is/Are’ (example: "Are dinosaurs big?").
        \item 3 points for the use of proper questioning words (example: "How big were the dinosaurs?").
    \end{itemize}  
\end{itemize}

\section{Statistical tests for investigating the relationship between curiosity and divergent QA performance}
\label{appendix:cur-div-qa}
\begin{figure}
\centering
\captionsetup{justification=centering,margin=1cm}
\includegraphics[width=1\linewidth]{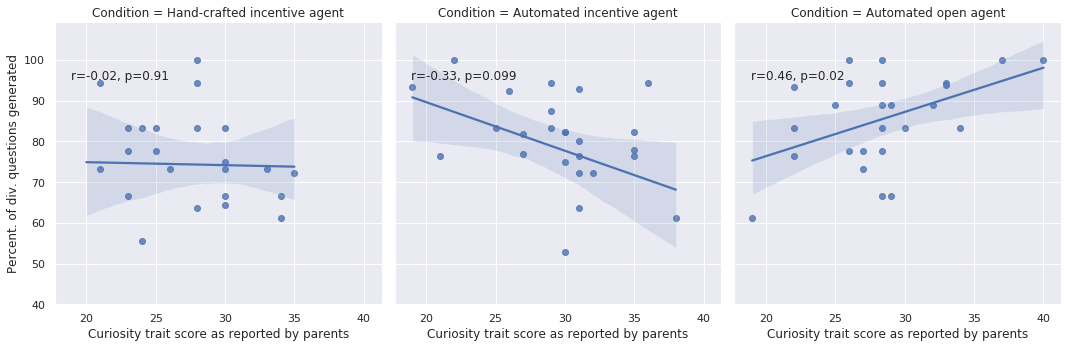}
\caption{\textbf{For the automated open agent, children's curiosity trait scores were strong predictors for their divergent question-asking abilities.}}
\label{fig:corr-qa-cur}
\end{figure}

In investigating the relationship between children's curiosity trait as reported by their parents and their divergent QA performance and the potential role of the experimental condition, we performed an ANCOVA test between the three groups, with the percentage of divergent questions generated during the training as a dependent variable and the curiosity trait score (as reported by the parents) as a co-variate. Results indicate a significant interaction between the divergent-QA performance and the curiosity scores, within the three conditions (F(2,71) = 4.06, p-value=0.02). Pairwise comparisons are then conducted in order to identify groups that are statistically different : we apply Bonferroni's multiple test correction. The post-hoc analysis show a statistically-significant difference only for the open agent condition (p-value=0.006). We also tested the significance of differences between the Pearson correlation coefficients for the three groups, by applying a Ronald Fisher z-transformation. As it can be seen in Figure~\ref{fig:corr-qa-cur} that the correlations were similar for the two incentive agents (z=1.7, p-value=0.14); however, the open agent led to a significantly different relationship (z=1.7, p-value=0.04) between the hand-crafted incentive agent and the automated open one and (z=2.8; p-value=0.002) between the automated incentive agent and the automated open one.

\end{appendices}
\bibliographystyle{abbrv}
\bibliography{bibfile}

\begin{thebibliography}{10}

\bibitem{Abdelghani2022}
R.~Abdelghani, P.-Y. Oudeyer, E.~Law, C.~de~Vulpillières, and H.~Sauzéon.
\newblock Conversational agents for training curiosity-driven learning in
  children.
\newblock {\em International Journal of Human-Computer studies}, 167, 2022.

\bibitem{alaimi2020}
M.~Alaimi, E.~Law, K.~D. Pantasdo, P.-Y. Oudeyer, and H.~Sauzeon.
\newblock Pedagogical agents for fostering question-asking skills in children.
\newblock In {\em Proceedings of the 2020 CHI Conference on Human Factors in
  Computing Systems}, CHI '20, page 1–13, New York, NY, USA, 2020.
  Association for Computing Machinery.

\bibitem{Aleven99}
V.~Aleven, K.~R. Koedinger, and K.~Cross.
\newblock Tutoring answer explanation fosters learning with understanding.
\newblock In {\em Artificial Intelligence in Education}, 1999.

\bibitem{bender2021}
E.~M. Bender, T.~Gebru, A.~McMillan-Major, and S.~Shmitchell.
\newblock On the dangers of stochastic parrots: Can language models be too
  big?��.
\newblock In {\em Proceedings of the 2021 ACM conference on fairness,
  accountability, and transparency}, pages 610--623, 2021.

\bibitem{berlyne1954}
D.~Berlyne.
\newblock A theory of human curiosity.
\newblock {\em British journal of psychology}, 45(3):180--191, August 1954.

\bibitem{bjork2017creating}
R.~Bjork.
\newblock Creating desirable difficulties to enhance learning.
\newblock {\em Carmarthen: Crown House Publishing}, 2017.

\bibitem{blodgett2021}
S.~L. Blodgett, G.~Lopez, A.~Olteanu, R.~Sim, and H.~Wallach.
\newblock Stereotyping norwegian salmon: An inventory of pitfalls in fairness
  benchmark datasets.
\newblock In {\em Proceedings of the 59th Annual Meeting of the Association for
  Computational Linguistics and the 11th International Joint Conference on
  Natural Language Processing (Volume 1: Long Papers)}, pages 1004--1015, 2021.

\bibitem{Brown2005}
J.~Brown, G.~A. Frishkoff, and M.~Esk{\'e}nazi.
\newblock Automatic question generation for vocabulary assessment.
\newblock In {\em HLT}, 2005.

\bibitem{Brown2020LanguageMA}
T.~B. Brown, B.~Mann, N.~Ryder, M.~Subbiah, J.~Kaplan, P.~Dhariwal,
  A.~Neelakantan, P.~Shyam, G.~Sastry, A.~Askell, S.~Agarwal, A.~Herbert-Voss,
  G.~Krueger, T.~J. Henighan, R.~Child, A.~Ramesh, D.~M. Ziegler, J.~Wu,
  C.~Winter, C.~Hesse, M.~Chen, E.~Sigler, M.~Litwin, S.~Gray, B.~Chess,
  J.~Clark, C.~Berner, S.~McCandlish, A.~Radford, I.~Sutskever, and D.~Amodei.
\newblock Language models are few-shot learners.
\newblock {\em ArXiv}, abs/2005.14165, 2020.

\bibitem{ceha2019a}
J.~Ceha, J.~Goh, C.~McDonald, D.~Kuli\'{c}, E.~Law, N.~Chhibber, and P.-Y.
  Oudeyer.
\newblock Expression of curiosity in social robots: Design, perception, and
  effects on behaviour.
\newblock {\em CHI Conference on Human Factors in Computing Systems (CHI2019)},
  page 1–12, 2019.

\bibitem{Cordova1996}
D.~Cordova and M.~Lepper.
\newblock Intrinsic motivation and the process of learning: Beneficial effects
  of contextualization, personalization, and choice.
\newblock {\em Journal of Educational Psychology}, 88:715--730, 12 1996.

\bibitem{devlin2019}
J.~Devlin, M.-W. Chang, K.~Lee, and K.~Toutanova.
\newblock {BERT}: Pre-training of deep bidirectional transformers for language
  understanding.
\newblock In {\em Proceedings of the 2019 Conference of the North {A}merican
  Chapter of the Association for Computational Linguistics: Human Language
  Technologies, Volume 1 (Long and Short Papers)}, pages 4171--4186,
  Minneapolis, Minnesota, June 2019. Association for Computational Linguistics.

\bibitem{diakidoy2002}
I.-A.~N. DIAKIDOY and G.~Spanoudis.
\newblock Domain specificity in creativity testing: A comparison of performance
  on a general divergent-thinking test and a parallel, content-specific test.
\newblock {\em The Journal of Creative Behavior}, 36(1):41--61, 2002.

\bibitem{gall1970}
M.~D. Gall.
\newblock The use of questions in teaching.
\newblock {\em Review of educational research}, 40(5):707--721, 1970.

\bibitem{Aschner1963}
J.~J. Gallagher and M.~J. Aschner.
\newblock A preliminary report on analyses of classroom interaction.
\newblock {\em Merrill-Palmer Quarterly}, 9:183--194, 1963.

\bibitem{goren2015}
G.~Gordon, C.~Breazeal, and S.~Engel.
\newblock Can children catch curiosity from a social robot?
\newblock {\em Proceedings of the 2015 ACM/IEEE International Conference on
  Human-Robot Interaction (HRI '15)}, 2015:91--98, 2015.

\bibitem{goupil2023}
L.~Goupil and J.~Proust.
\newblock Curiosity as a metacognitive feeling.
\newblock {\em Cognition}, 231:105325, 2023.

\bibitem{graesser1992}
A.~C. Graesser, N.~Person, and J.~Huber.
\newblock Mechanisms that generate questions.
\newblock In T.~W. Lauer, P.~E., and A.~C. Graesser, editors, {\em Questions
  and Information Systems}, pages 167--187. Lawrence Erlbaum Associates,
  Mahwah, NJ, 1992.

\bibitem{Graesser1994}
A.~C. Graesser and N.~K. Person.
\newblock Question asking during tutoring.
\newblock {\em American Educational Research Journal}, 31:104--137, 1994.

\bibitem{Guay2022}
F.~Guay.
\newblock Applying self-determination theory to education: Regulations types,
  psychological needs, and autonomy supporting behaviors.
\newblock {\em Canadian Journal of School Psychology}, 37(1):75--92, 2022.

\bibitem{Nasa-tlx}
S.~G. Hart and L.~E. Staveland.
\newblock Development of nasa-tlx (task load index): Results of empirical and
  theoretical research.
\newblock {\em Advances in psychology}, 52:139--183, 1988.

\bibitem{https://doi.org/10.48550/arxiv.2106.03164}
R.~He, L.~Liu, H.~Ye, Q.~Tan, B.~Ding, L.~Cheng, J.-W. Low, L.~Bing, and L.~Si.
\newblock On the effectiveness of adapter-based tuning for pretrained language
  model adaptation, 2021.

\bibitem{Humphries2015}
J.~Humphries and M.~Ness.
\newblock Beyond who, what, where, when, why, and how: Preparing students to
  generate questions in the age of common core standards.
\newblock {\em Journal ofResearch in Childhood Education}, 29:551--561, 2015.

\bibitem{10.3389/fnbeh.2012.00005}
M.~Jepma, R.~Verdonschot, H.~van Steenbergen, S.~Rombouts, and S.~Nieuwenhuis.
\newblock Neural mechanisms underlying the induction and relief of perceptual
  curiosity.
\newblock {\em Frontiers in Behavioral Neuroscience}, 6, 2012.

\bibitem{jiang-etal-2021-know}
Z.~Jiang, J.~Araki, H.~Ding, and G.~Neubig.
\newblock How can we know when language models know? on the calibration of
  language models for question answering.
\newblock {\em Transactions of the Association for Computational Linguistics},
  9:962--977, 2021.

\bibitem{jirout2018}
J.~Jirout, V.~Vitiello, and S.~Zumbrunn.
\newblock {\em Curiosity in schools}, pages 243--266.
\newblock 07 2018.

\bibitem{Jones2018}
A.~Jones, S.~Bull, and G.~Castellano.
\newblock “i know that now, i’m going to learn this next” promoting
  self-regulated learning with a robotic tutor.
\newblock {\em International Journal of Social Robotics}, 10:439--454, 2018.

\bibitem{john1998}
J.~T. Jost, A.~W. Kruglanski, and T.~O. Nelson.
\newblock Social metacognition: An expansionist review.
\newblock {\em Personality and Social Psychology Review}, 2(2):137--154, 1998.
\newblock PMID: 15647141.

\bibitem{Murayama2019}
M.~K., , F.~L., and S.~M.
\newblock Process account of curiosity and interest: A reward-learning
  perspective.
\newblock {\em Educational Psychology Review}, 31:875--895, Dec. 2019.

\bibitem{kang2009}
M.~Kang, M.~Hsu, I.~Krajbich, G.~Loewenstein, S.~M. McClure, J.~Wang, and
  C.~Camerer.
\newblock The wick in the candle of learning: Epistemic curiosity activates
  reward circuitry and enhances memory.
\newblock {\em Psychological science}, 20(8):963--973, August 2009.

\bibitem{kashdan2004}
T.~Kashdan, P.~Rose, and F.~Fincham.
\newblock Curiosity and exploration: Facilitating positive subjective
  experiences and personal growth opportunities.
\newblock {\em Journal of Personality Assessment}, 82(3):291--305, June 2004.

\bibitem{kasneci2023}
E.~Kasneci, K.~Se{\ss}ler, S.~K{\"u}chemann, M.~Bannert, D.~Dementieva,
  F.~Fischer, U.~Gasser, G.~Groh, S.~G{\"u}nnemann, E.~H{\"u}llermeier, et~al.
\newblock Chatgpt for good? on opportunities and challenges of large language
  models for education.
\newblock {\em Learning and Individual Differences}, 103:102274, 2023.

\bibitem{kumar2021}
S.~Kumar and P.~P. Talukdar.
\newblock Reordering examples helps during priming-based few-shot learning.
\newblock {\em CoRR}, abs/2106.01751, 2021.

\bibitem{law2016}
E.~Law, M.~Yin, J.~Goh, K.~Chen, M.~A. Terry, and K.~Z. Gajos.
\newblock Curiosity killed the cat, but makes crowdwork better.
\newblock In {\em Proceedings of the 2016 ACM CHI Conference on Human Factors
  in Computing Systems (CHI '16)}, pages 4098--4110, 2016.

\bibitem{li-liang-2021-prefix}
X.~L. Li and P.~Liang.
\newblock Prefix-tuning: Optimizing continuous prompts for generation.
\newblock In {\em Proceedings of the 59th Annual Meeting of the Association for
  Computational Linguistics and the 11th International Joint Conference on
  Natural Language Processing (Volume 1: Long Papers)}, pages 4582--4597,
  Online, Aug. 2021. Association for Computational Linguistics.

\bibitem{litman2005}
J.~Litman.
\newblock Curiosity and the pleasures of learning: Wanting and liking new
  information.
\newblock {\em Cognition \& Emotion}, 19(6):793--814, September 2005.

\bibitem{liu2022makes}
J.~Liu, D.~Shen, Y.~Zhang, B.~Dolan, L.~Carin, and W.~Chen.
\newblock What makes good in-context examples for {GPT}-3?
\newblock In {\em Proceedings of Deep Learning Inside Out (DeeLIO 2022): The
  3rd Workshop on Knowledge Extraction and Integration for Deep Learning
  Architectures}, pages 100--114, Dublin, Ireland and Online, May 2022.
  Association for Computational Linguistics.

\bibitem{loewenstein1994}
G.~Loewenstein.
\newblock {The Psychology of Curiosity: A Review and Reinterpretation}.
\newblock {\em Psychological Bulletin}, 116(1):75--98, 1994.

\bibitem{lu2021fantastically}
Y.~Lu, M.~Bartolo, A.~Moore, S.~Riedel, and P.~Stenetorp.
\newblock Fantastically ordered prompts and where to find them: Overcoming
  few-shot prompt order sensitivity.
\newblock In {\em Proceedings of the 60th Annual Meeting of the Association for
  Computational Linguistics (Volume 1: Long Papers)}, pages 8086--8098, Dublin,
  Ireland, May 2022. Association for Computational Linguistics.

\bibitem{marx1999}
A.~Marx, U.~Fuhrer, and T.~Hartig.
\newblock Effects of classroom seating arrangements on children's
  question-asking.
\newblock {\em Learning Environments Research}, 2:249--263, 1999.

\bibitem{mehta2018}
H.~Mehta, R.~Dubey, and T.~Lombrozo.
\newblock Your liking is my curiosity: a social popularity intervention to
  induce curiosity.
\newblock In {\em Proceedings of the 40th Annual Meeting of the Cognitive
  Science Society}, pages 756--761, 2018.

\bibitem{METCALFE202040}
J.~Metcalfe, B.~L. Schwartz, and T.~S. Eich.
\newblock Epistemic curiosity and the region of proximal learning.
\newblock {\em Current Opinion in Behavioral Sciences}, 35:40--47, 2020.
\newblock Curiosity (Explore vs Exploit).

\bibitem{min2022rethinking}
S.~Min, X.~Lyu, A.~Holtzman, M.~Artetxe, M.~Lewis, H.~Hajishirzi, and
  L.~Zettlemoyer.
\newblock Rethinking the role of demonstrations: What makes in-context learning
  work?
\newblock {\em arXiv preprint arXiv:2202.12837}, 2022.

\bibitem{oudeyer2008}
P.-Y. Oudeyer, F.~Kaplan, and V.~V. Hafner.
\newblock Intrinsic motivation systems for autonomous mental development.
\newblock {\em IEEE Transactions on Evolutionary Computation}, 11(2):265--286,
  2007.

\bibitem{Post2015}
T.~Post and J.~H. Walma van~der Molen.
\newblock Development and validation of a questionnaire to measure primary
  school children’s images of and attitudes towards curiosity (the ciac
  questionnaire).
\newblock {\em Motivation and Emotion}, 42,1:159--178, 2015.

\bibitem{Raffel2020}
C.~Raffel, N.~Shazeer, A.~Roberts, K.~Lee, S.~Narang, M.~Matena, Y.~Zhou,
  W.~Li, and P.~J. Liu.
\newblock Exploring the limits of transfer learning with a unified text-to-text
  transformer.
\newblock {\em Journal of Machine Learning Research}, 21(140):1--67, 2020.

\bibitem{reio2000}
T.~Reio and A.~Wiswell.
\newblock Field investigation of the relationship among adult curiosity,
  workplace learning, and job performance.
\newblock {\em Human Resource Development Quarterly}, 11(1):5--30, March 2000.

\bibitem{Roebers2007}
C.~Roebers, N.~von~der Linden, and P.~Howie.
\newblock Favourable and unfavourable conditions for children's confidence
  judgments.
\newblock {\em British Journal of Developmental Psychology}, 25:109--134, 2007.

\bibitem{rubin2021learning}
O.~Rubin, J.~Herzig, and J.~Berant.
\newblock Learning to retrieve prompts for in-context learning.
\newblock {\em arXiv preprint arXiv:2112.08633}, 2021.

\bibitem{ryan2000selfdetermination}
R.~M. Ryan and E.~L. Deci.
\newblock Self-determination theory and the facilitation of intrinsic
  motivation, social development, and well-being, 2000.

\bibitem{Bereiter1992}
M.~Scardamalia and C.~Bereiter.
\newblock Text-based and knowledge based questioning by children.
\newblock {\em Cognition and Instruction}, 9:177--199, 1992.

\bibitem{prachi2018}
P.~Shah, H.~Weeks, B.~Richards, and N.~Kaciroti.
\newblock Early childhood curiosity and kindergarten reading and math academic
  achievement.
\newblock {\em Pediatric Research}, 84:380--386, April 2018.

\bibitem{Silvervarg2010}
A.~Silvervarg Flycht-Eriksson and A.~Jönsson.
\newblock Towards a conversational pedagogical agent capable of affecting
  attitudes and self-efficacy.
\newblock In {\em Proceedings of the Second Workshop on Natural Language
  Processing in Support of Learning: Metrics, Feedback and Connectivity}, 2010.

\bibitem{stahl15}
A.~Stahl and L.~Feigenson.
\newblock Observing the unexpected enhances infants' learning and exploration.
\newblock {\em Science}, 348(6230):91--94, April 2015.

\bibitem{Steuer2021}
T.~Steuer, A.~Filighera, T.~Meuser, and C.~Rensing.
\newblock I do not understand what i cannot define: Automatic question
  generation with pedagogically-driven content selection.
\newblock {\em ArXiv}, abs/2110.04123, 2021.

\bibitem{article}
S.~Stumm, B.~Hell, and T.~Chamorro-Premuzic.
\newblock The hungry mind -- intellectual curiosity is the third pillar of
  academic performance.
\newblock {\em Perspectives on Psychological Science}, 6:574--588, 10 2011.

\bibitem{Sultan2014}
M.~A. Sultan, S.~Bethard, and T.~Sumner.
\newblock Towards automatic identification of core concepts in educational
  resources.
\newblock {\em Proceedings of the ACM/IEEE Joint Conference on Digital
  Libraries}, pages 379--388, 12 2014.

\bibitem{Ten2021}
A.~Ten, P.~Kaushik, P.-Y. Oudeyer, and J.~Gottlieb.
\newblock Humans monitor learning progress in curiosity-driven exploration.
\newblock {\em Nature Communications}, 12:5972, 10 2021.

\bibitem{tullis2015}
J.~G. Tullis and A.~S. Benjamin.
\newblock Cueing others’ memories.
\newblock {\em Memory \& cognition}, 143(4):634--646, 2015.

\bibitem{vallerand1989}
R.~J. Vallerand, M.~R. Blais, N.~M. Brière, and L.~G. Pelletier.
\newblock Construction et validation de l'échelle de motivation en éducation
  (eme) [construction and validation of the motivation toward education scale].
\newblock {\em Canadian Journal of Behavioural Science / Revue canadienne des
  sciences du comportement}, 21:323--349, 1989.

\bibitem{wilen1991}
W.~W. Wilen.
\newblock Questioning skills, for teachers. what research says to the teacher.
\newblock 1991.

\bibitem{Xingdi2022}
X.~Yuan, T.~Wang, Y.-H. Wang, E.~Fine, R.~Abdelghani, P.~Lucas, H.~Sauzéon,
  and P.-Y. Oudeyer.
\newblock Selecting better samples from pre-trained llms: A case study on
  question generation, 2022.

\bibitem{zhao2021calibrate}
Z.~Zhao, E.~Wallace, S.~Feng, D.~Klein, and S.~Singh.
\newblock Calibrate before use: Improving few-shot performance of language
  models.
\newblock In {\em International Conference on Machine Learning}, pages
  12697--12706. PMLR, 2021.

\end{thebibliography}
\end{document}